\documentclass[10pt,twocolumn,letterpaper]{article}

\usepackage{cvpr}
\usepackage{times}
\usepackage{epsfig}
\usepackage{graphicx}
\usepackage{amsmath}
\usepackage{amssymb}

\usepackage{booktabs,multirow}
\usepackage{makecell}
\usepackage{caption}
\usepackage{subcaption}
\usepackage{algorithm}
\usepackage{algpseudocode}
\usepackage{algorithmicx}
\usepackage{float}
\usepackage{url}

\usepackage{color}
\usepackage{xcolor,colortbl}
\usepackage{framed}
\usepackage{enumitem}
\usepackage{booktabs}

\definecolor{shadecolor}{rgb}{0.9,0.9,0.9}
\definecolor{Gray}{gray}{0.9}

\algnewcommand\algorithmicinput{\textbf{Input:}}
\algnewcommand\INPUT{\item[\algorithmicinput]}

\newcommand{\tabincell}[2]{\begin{tabular}{@{}#1@{}}#2\end{tabular}}

\usepackage[pagebackref=true,breaklinks=true,letterpaper=true,colorlinks,bookmarks=false]{hyperref}

\cvprfinalcopy 

\def\cvprPaperID{556}

\ifcvprfinal\fi

\begin{document}

\title{Open Compound Domain Adaptation}

\author{Ziwei Liu$^{1}$\thanks{Equal contribution.} ~~~~~ Zhongqi Miao$^{2*}$ ~~~~~ Xingang Pan$^{1}$ ~~~~~ Xiaohang Zhan$^{1}$ \\ Dahua Lin$^{1}$ ~~~~~ Stella X. Yu$^{2}$ ~~~~~ Boqing Gong$^{3}$ \\
$^{1}$ The Chinese University of Hong Kong ~~~~~~~ $^{2}$ UC Berkeley / ICSI ~~~~~~~ $^{3}$ Google Inc. \\
{\tt\small \url{https://liuziwei7.github.io/projects/CompoundDomain.html}}
\vspace{-12pt}
}

\maketitle
\thispagestyle{empty}

\begin{abstract}

A typical domain adaptation approach is to adapt models trained on the annotated data in a source domain (e.g., sunny weather) for achieving high performance on the test data in a target domain (e.g., rainy weather).  Whether the target contains a single homogeneous domain or multiple heterogeneous domains, existing works always assume that there exist clear distinctions between the domains, which is often not true in practice (e.g., changes in weather). We study an open compound domain adaptation (OCDA) problem, in which the target is a compound of multiple homogeneous domains without domain labels, reflecting realistic data collection from mixed and novel situations. We propose a new approach based on two technical insights into OCDA: 1) a curriculum domain adaptation strategy to bootstrap generalization across domains in a data-driven self-organizing fashion and 2) a memory module to increase the model's agility towards novel domains. Our experiments on digit classification, facial expression recognition, semantic segmentation, and reinforcement learning demonstrate the effectiveness of our approach.

\end{abstract}

\begin{figure}[t]
  \centering
  \includegraphics[width=0.48\textwidth,clip]{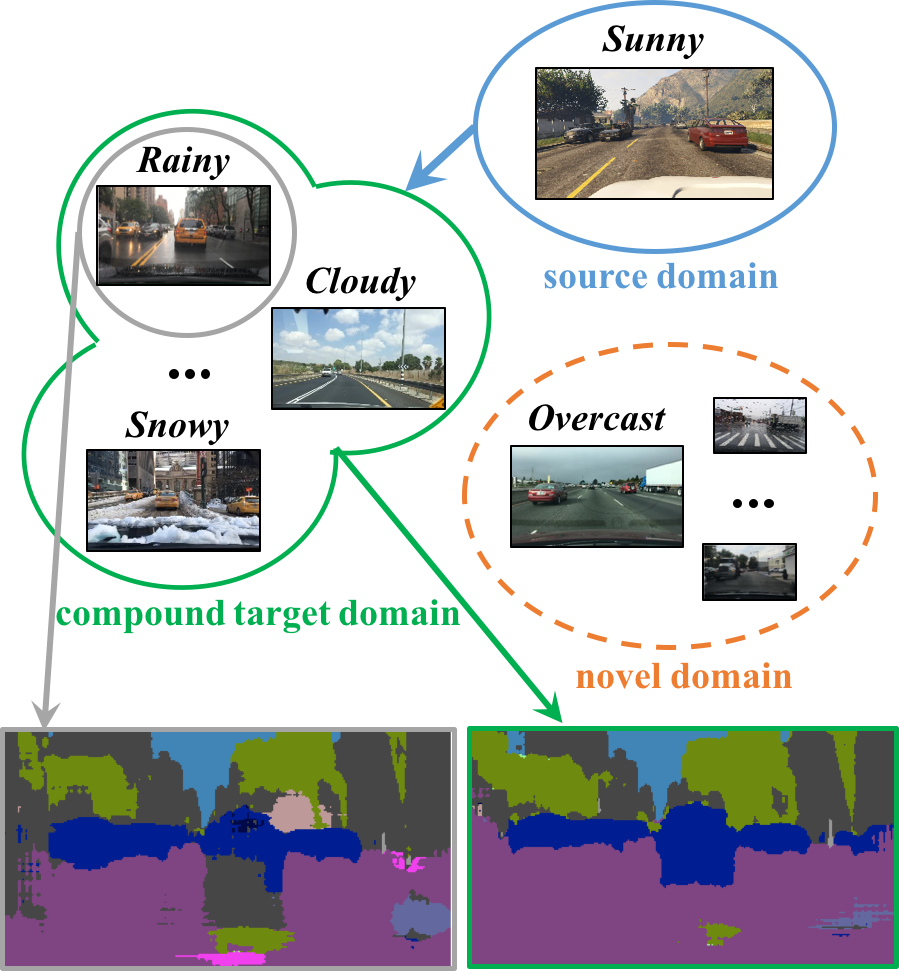}
  \caption{\textbf{Open compound domain adaptation.}  Unlike existing domain adaptation which assumes clear distinctions between discrete domains (cf.\ the examples in gray frames),  our compound target domain is a combination of multiple traditionally homogeneous domains without any  domain labels. We also allow novel domains to show up at the inference time.}
  \label{fig:problem}
\end{figure}

\begin{table*}[t]
    \centering
    \caption{\textbf{Comparison of domain adaptation settings.}  
    {Domain Labels} tell to which domain each instance belongs.  
    {Open Classes} refer to novel classes showing up during testing but not training.
    {Open Domains} are the domains of which no  instances are seen during training.
    }
    \label{tab:comparison}
    \vspace{-5pt}
    \begin{tabular}{l|c|c|c|c} 
    \Xhline{1pt}
    {\bf Domain Adaptation Setting} & {\bf \# Target Domains} & {\bf Domain Labels} & {\bf Open Classes} & {\bf Open Domains} \\ \Xhline{0.8pt}
    Unsupervised Domain Adaptation & single & known & $\times$ & $\times$ \\ \hline
    Multi-Target Domain Adaptation & multiple & known & $\times$ & $\times$ \\ \hline
    Open/Partial Set Domain Adaptation & single & known & $\checkmark$ & $\times$ \\ \hline
    \bf Open Compound Domain Adaptation & \bf multiple &\bf unknown & \bf $\times$ & \bf $\checkmark$ \\ \Xhline{1pt}
    \end{tabular}
    \vspace{-10pt}
\end{table*}

\section{Introduction}

Supervised learning  can achieve competitive performance for a visual task when the test data is drawn from the same underlying distribution as the training data.  This assumption, unfortunately, often does not hold in reality, \eg, the test data may contain the same class of objects as the training data but  different backgrounds, poses, and appearances~\cite{saenko2010adapting, torralba2011}.  

The goal of domain adaptation is to adapt the model learned on the training data to the test data of a different distribution ~\cite{saenko2010adapting, pan2010domain, gong2012geodesic}.
Such a distributional gap is often formulated as a shift between discrete concepts of well defined data domains, \eg, images collected in sunny weather versus those in rainy weather.  
Though domain generalization~\cite{li2017domain, li2018learning} and latent domain adaptation~\cite{hoffman2012discovering, gong2013reshaping} have attempted to tackle complex target domains, most existing works usually assume that there is a known clear distinction between domains ~\cite{gong2012geodesic,ganin2016domain, tzeng2017adversarial, long2017deep, saito2018maximum}.

Such a known and clear distinction between domains is hard to define in practice, \eg, test images could be collected in mixed, continually varying, and sometimes never seen weather conditions.  With numerous factors jointly contributing to data variance, it becomes implausible to separate data into discrete domains. 

We propose to study {\it open compound domain adaptation} (OCDA),  a continuous and more realistic setting for domain adaptation (cf.~Figure~\ref{fig:problem} and Table~\ref{tab:comparison}).
The task is to learn a model from  labeled {\it source domain} data and adapt it to unlabeled {\it compound target domain} data which could differ from the source domain on various factors.  Our target domain can be regarded as a combination of multiple traditionally homogeneous domains where each is distinctive on one or two major factors, and yet none of the domain labels are given. For example, the five well-known datasets on digits recognition (SVHN~\cite{Yuval2011svhn}, MNIST~\cite{lecun1998gradient}, MNIST-M~\cite{ganin2015unsupervised}, USPS~\cite{Hull1994usps}, and SynNum~\cite{ganin2015unsupervised}) mainly differ from each other by the backgrounds and text fonts. It is not necessarily the best practice, and not feasible under some scenarios, to consider them as distinct domains. Instead, our compound target domain pools them together. Furthermore, at the inference stage, OCDA tests the model not only in the compound target domain but also in open domains that have previously unseen during training.

In our OCDA setting, the target domain no longer has a  predominantly uni-modal distribution, posing challenges to existing domain adaptation methods. We propose a novel approach based on two technical insights into OCDA: 1) a curriculum domain adaptation strategy to bootstrap generalization across domain distinction in a data-driven self-organizing fashion and 2) a memory module to increase the model's agility towards novel domains.

Unlike existing curriculum adaptation methods ~\cite{zhang2019curriculum,SynRealDataFog19,liu2016fashion,li2017not,zou2018unsupervised,zou2019confidence} that rely on some holistic measure of instance difficulty,
we schedule the learning of unlabeled instances in the compound target domain according to their {\it  individual gaps} to the labeled source domain, so that we solve an incrementally harder domain adaptation problem till we cover the entire target domain.

Specifically, we first train a neural network to 1) discriminate between classes in the labeled source domain and to 2) capture domain invariance from the easy target instances which differ the least from labeled source domain data.  Once the network can no longer differentiate between the source domain and the easy target domain data, we feed the network harder target instances, which are further away from the source domain.  The network learns to remain discriminative to the classification task and yet grow more robust to the entire compound target domain.

Technically, we must address the challenge of characterizing each instance's gap to the source domain.  We first extract domain-specific feature representations from the data and then rank the target instances according to their distances to the source domain in that feature space, assuming that such features do not contribute to and even distract the network from learning  discriminative features for classification.  We use a class-confusion loss to distill the domain-specific factors and formulate it as a conventional cross-entropy loss with a randomized class label twist. 

Our second technical insight is to prepare our model for open domains during inference with a memory module that effectively augments the representations of an input for classification. Intuitively, if the input is close enough to the source domain, the feature extracted from itself can most likely already result in accurate classification.  Otherwise, the input-activated memory features can step in and play a more important role.  Consequently, this memory-augmented network is more agile at handling open domains than its vanilla counterpart.

To summarize, we make the following contributions. {\bf 1)} We  extend the traditional discrete domain adaptation to OCDA, a more realistic continuous domain adaptation setting.  {\bf 2)} We develop an OCDA solution with two key technical insights: instance-specific curriculum domain adaptation for handling the target of mixed domains  and memory augmented features  for handling open domains. {\bf 3)} We design several benchmarks on classification, recognition, segmentation, and reinforcement learning, and conduct comprehensive experiments to evaluate our approach under the OCDA setting.


\begin{figure*}[t]
  \centering
  \includegraphics[width=\textwidth]{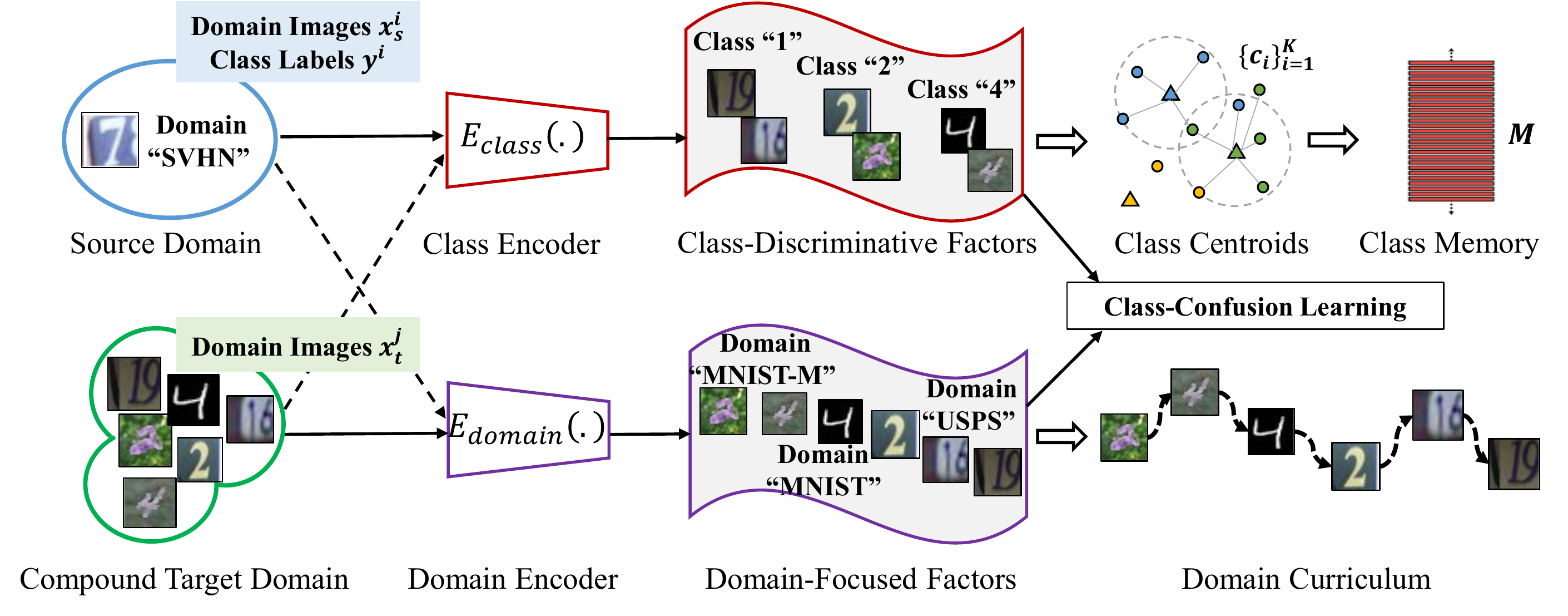}
  \caption{\textbf{Overview of disentangling domain characteristics and curriculum domain adaptation.} We separate characteristics specific to domains from those discriminative between classes.  It is achieved by a class-confusion algorithm in an unsupervised manner.  The teased out domain feature is used to construct a curriculum for domain-robust learning.}
  \label{fig:pipeline_disentangle}
\end{figure*}


\section{Related Works}


We review literature according to Table~\ref{tab:comparison}.

\vspace{2pt}
\noindent
\textbf{Unsupervised Domain Adaptation.}  The goal is to retain recognition accuracies in new domains without ground truth annotations  ~\cite{saenko2010adapting, torralba2011, venkateswara2017deep, peng2019moment}.  Representative techniques include latent distribution alignment~\cite{gong2012geodesic}, back-propagation~\cite{ganin2015unsupervised}, gradient reversal~\cite{ganin2016domain}, adversarial discrimination~\cite{tzeng2017adversarial}, joint maximum mean discrepancy~\cite{long2017deep}, cycle consistency~\cite{hoffman2017cycada} and maximum classifier discrepancy~\cite{saito2018maximum}.  While their results are promising, this traditional domain adaptation setting focuses on ``one source domain, one target domain'', and cannot deal with more complicated scenarios where multiple target domains are present.

\vspace{2pt}
\noindent
\textbf{Latent \& Multi-Target Domain Adaptation.}
The goal is to extend unsupervised domain adaptation to latent~\cite{hoffman2012discovering, xiong2014latent, mancini2019inferring} or multiple  \cite{gong2013reshaping, gholami2018unsupervised, yu2018multi} or continuous~\cite{Bobu2018adapting, gong2018dlow, Mancini_2019_CVPR, wu2019ace} target domains, when only the source domain has class labels.  These methods usually assume clear domain distinction or require domain labels (\eg test instance $i$ belongs to the target domain $j$), but this assumption rarely holds in the real-world scenario.  Here we take one step further towards compound domain adaptation, where both category labels and domain labels in the test set are unavailable.

\vspace{2pt}
\noindent
\textbf{Open/Partial Set Domain Adaptation.}
Another route of research aims to tackle the category sharing/unsharing issues between source and target domain, namely open set~\cite{panareda2017open, saito2018open} and partial set~\cite{zhang2018importance, cao2018partial} domain adaptation. They assume that the target domain contains either 1) new categories that don't appear in source domain; or 2) only a subset of categories that appear in source domain. Both settings concern the ``openness'' of categories. Instead, here we investigate the ``openness'' of domains, \ie there are novel domains existing that are absent in the training phase.

\vspace{2pt}
\noindent
\textbf{Domain Generalized/Agnostic Learning.}
Domain generalization~\cite{xu2014exploiting, li2018domain, li2018learning} and domain agnostic learning~\cite{peng2019domain, Chen_2019_CVPR} aim to learn universal representations that can be applied in a domain-invariant manner. Since these methods focus on learning semantic representations that are invariant to the domain shift, they largely neglect the latent structures inside the target domains. In this work, we explicitly model the latent structures inside the compound target domain by leveraging the learned domain-focused factors for curriculum scheduling and dynamic adaptation.

\section{Our Approach to OCDA}

\begin{figure*}[t]
  \centering
  \includegraphics[width=\textwidth]{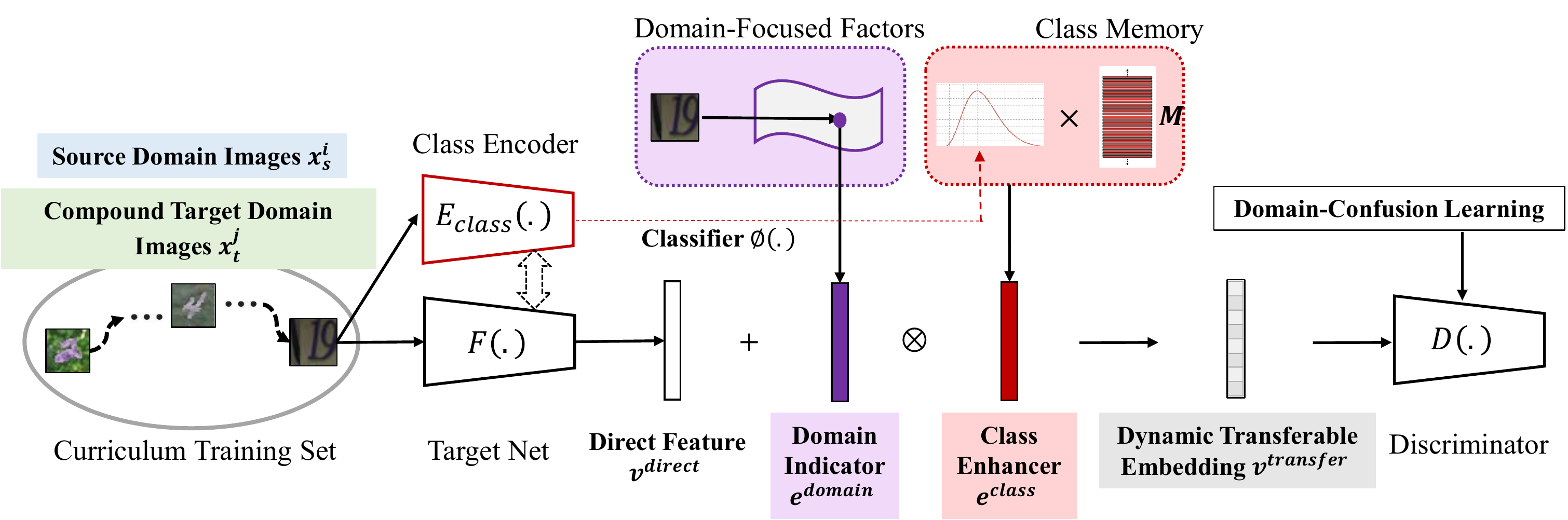}
  \caption{\textbf{Overview of the memory-enhanced deep neural network.} We enhance our network with a memory module that facilitates knowledge transfer from the source domain to target domain instances, so that the network can dynamically balance the input information and the memory-transferred knowledge for more agility towards previously unseen domains.}
  \label{fig:pipeline_memory}
  \vspace{-10pt}
\end{figure*}

Figures~\ref{fig:pipeline_disentangle} and~\ref{fig:pipeline_memory} present our overall workflows.  There are three major components: 1) disentangling domain characteristics with only class labels in the source domain,
2) scheduling data for curriculum domain adaptation, and 3) a memory module for handling new domains.

\subsection{Disentangling Domain Characteristics}
We separate characteristics specific to domains from those discriminative between classes.  They allow us to construct a curriculum for increment domain adaptation.

We first train a neural network classifier using the labeled source domain data $\{x^{i}, y^{i}\}_{i}$. Let $E_{class}(\cdot)$ denote the encoder up to the second-to-the-last layer and $\Phi(E_{class}(\cdot))$ the classifier.  The encoder captures primarily the class-discriminative representation of the data.

We assume that all the factors not covered by this class-discriminative encoder reflect domain characteristics. 
They can be extracted by another encoder $E_{domain}(\cdot)$ that satisfies two properties:  {\bf 1)} Completeness: $Decoder(E_{class}(x), E_{domain}(x))\approx x$, i.e., the outputs of the two encoders shall provide sufficient information for a decoder to reconstruct the input, and  {\bf 2)} Orthogonality:  the domain encoder $E_{domain}(x)$ shall have little mutual information with the class encoder $E_{class}(x)$.  We leave the algorithmic details for meeting the first property to the appendices as they are not our novelty.

For the orthogonality between $E_{domain}(x)$ and  $E_{class}(x)$, we propose a \textbf{class-confusion algorithm}, which alternates between the two sub-problems below: 
\begin{align}
    &\min_{E_{domain}}\; && -\sum_i z_{random}^i \log D(E_{domain}(x^i)), \\
    &\min_D\; && -\sum_i y^i\log D(E_{domain}(x^i)),
\end{align}
where superscript $i$ is the instance index, and $D(\cdot)$ is a discriminator the domain-encoder $E_{domain}(\cdot)$ tries to confuse. We first train the discriminator $D(\cdot)$ with the labeled data in the source domain.  For the data in the target domain, we assign them pseudo-labels by the classifier $\Phi(E_{class}(\cdot))$ we have trained earlier. The learned domain encoder $E_{domain}(\cdot)$ is class-confusing  due to $z_{random}^i$, a random label uniformly chosen in the label space.  As the classifier $D(\cdot)$ is trained, the first sub-problem essentially learns the domain-encoder such that it classifies the input $x^i$ into a random class $z_{random}^i$.  Algorithm~\ref{alg:disentangle} details our domain disentanglement process. 

Figure~\ref{fig:manifold} (a) and (b) visualize the examples embedded by the class encoder $E_{class}(\cdot)$ and domain encoder $E_{domain}(\cdot)$, respectively.  The class encoder places instances in the same class in a cluster, while the domain encoder places instances according to their common appearances,  regardless of their classes.

\begin{algorithm}[ht]
    \caption{Domain Disentanglement.}
    \label{alg:disentangle}
    \begin{algorithmic}
    \INPUT{The class encoder $E_{class}(\cdot)$ and classifier $\Phi$ have been trained using source-domain data, $Deccoder(\cdot)$: the decoder, $C$: the number of classes, $\gamma$: a constant.}
    \For{k iterations}
        \State Sample mini-batch $\{x^i\}$.
        \State Compute pseudo labels $y_{pseudo}^i \gets \Phi\left(E_{class}\left(x^i\right)\right)$.
        \State Update the discriminator $D$.
        \State Prepare random labels $z_{random}^i \sim uniform\{0,1,...,C-1\}$.
        \State Compute adversarial loss: $L_{adv} \gets \sum_i -z_{random}^i\log\left(D\left(E_{domain}(x^i)\right)\right)$.
        \State Compute reconstruction loss: $L_{rec} \gets \sum_i \| Decoder\left(E_{class}\left(x^i\right), E_{domain}\left(x^i\right)\right) - x^i\|_2$.
        \State Update the domain encoder $E_{domain}$ with: $\nabla_{\theta_{E_{domain}}} \left(L_{adv} + \gamma L_{rec}\right)$.
    \EndFor
    \end{algorithmic}
\end{algorithm}

\subsection{Curriculum Domain Adaptation}
We rank all the instances in the compound target domain according to their distances to the source domain, to be used for curriculum domain adaptation~\cite{zhang2019curriculum}. We compute the {\it domain gap} between a target instance $x_t$ and the source domain $\{x_s^{m}\}$ as their mean distance
in the domain feature space: $\texttt{mean}_{m}(\|E_{domain}(x_t)-E_{domain}(x_s^m)\|_2)$.

\begin{figure*}[t]
  \centering
  \includegraphics[width=1.0\textwidth]{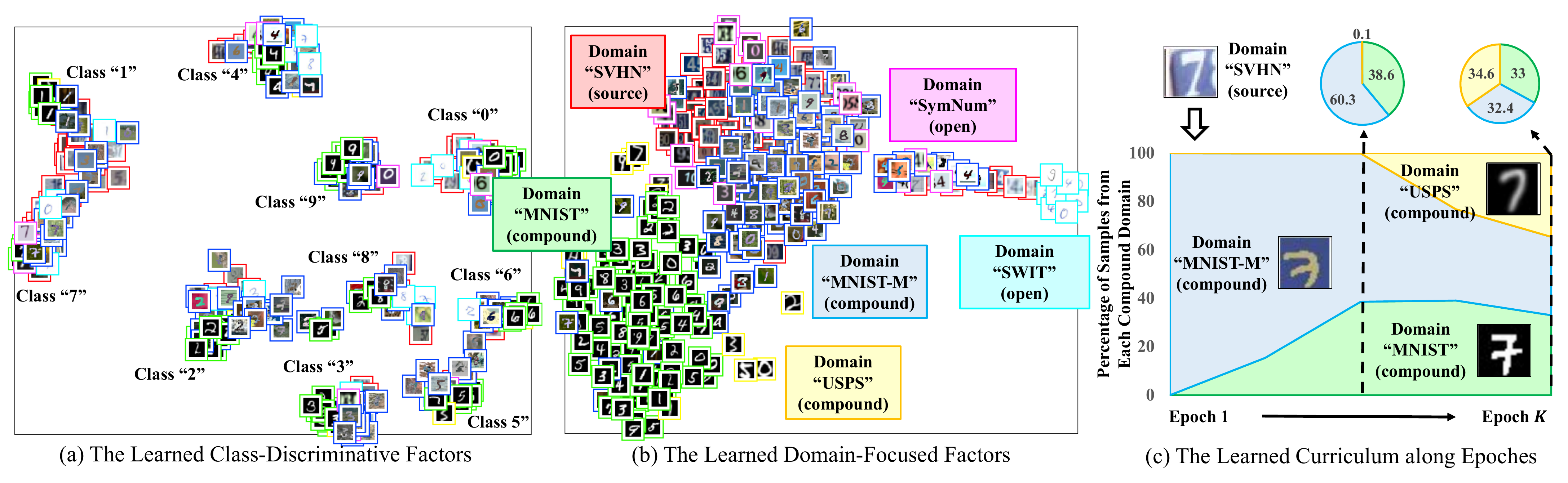}
  \caption{\textbf{t-SNE Visualization} of our (a) class-discriminative features, (b) domain features, and (c) curriculum. Our framework disentangles the mixed-domain data into class-discriminative factors and domain-focused factors. We use the domain-focused factors to construct a learning curriculum for domain adaptation.}
  \label{fig:manifold}
\end{figure*}

We train the network in stages, a few epochs at a time, gradually recruiting more instances that are increasingly far from the source domain.  At each stage of the curriculum learning, we minimize two losses: One is the cross-entropy loss defined over the labeled source domain, and the other is the {domain-confusion loss}~\cite{tzeng2017adversarial} computed between the source domain and the currently covered target instances.  Figure~\ref{fig:manifold} (c) illustrates a curriculum in our experiments.


\subsection{Memory Module for Open Domains}

Existing domain adaptation methods often use the features $v_{direct}$ extracted directly from the input for adaptation.   When the input comes from a new domain that  significantly differs from the seen domains during training, this  representation becomes inadequate and could fool the classifier. We propose a memory module to enhance our model; It allows knowledge transfer from the source domain so that the network can dynamically balance the input-conveyed information and the memory-transferred knowledge for more classification agility towards previously unseen domains.


\vspace{2pt}
\noindent
\textbf{Class Memory $M$.}
We design a memory module $M$ to store the class information from the source domain.  Inspired by ~\cite{snell2017prototypical, Pan_2019_CVPR, liu2019large} on prototype analysis, we also use class centroids $\{c_{k}\}_{k=1}^{K}$ to construct our memory $M$, where $K$ is the number of object classes.

\vspace{2pt}
\noindent
\textbf{Enhancer $v_{enhance}$.}
For each input instance, we build an enhancer to augment its direct representation $v_{direct}$ with knowledge in the memory about the source domain: $v_{enhance} = (\Psi(v_{direct}))^T M = \sum_{k=1}^K \psi_{k}c_k$,
where $\Psi(\cdot)$ is a softmax function.  We add this enhancer to the direct representation $v_{direct}$, weighted by a domain indicator. 

\vspace{2pt}
\noindent
\textbf{Domain Indicator $e_{domain}$.}
With open domains, the network must dynamically calibrate how much knowledge to transfer from the source domain and how much to rely on the direct representation $v_{direct}$ of the input.  Intuitively, the larger domain gap between an input $x$ and the source domain, the more weight on the memory feature.  We design a domain indicator for such domain awareness: $e_{domain} = T(E_{domain}(x))$,
where $T(\cdot)$ is a lightweight network with the \texttt{tanh} activation functions and $E_{domain}(\cdot)$ is the domain encoder we have learned earlier.

\vspace{2pt}
\noindent
\textbf{Source-Enhanced Representation $v_{transfer}$.}
Our final representation of the input is a dynamically
balanced version between the direct image feature and the memory enhanced feature:
\begin{equation}
v_{transfer} = v_{direct} + e_{domain} \otimes v_{enhance},
\end{equation}
which transfers class-discriminative knowledge from the labeled source domain to the input in a domain-aware manner.  Operator $\otimes$ is element-wise multiplication.  Adopting cosine classifiers~\cite{liu2017sphereface, gidaris2018dynamic}, we $\ell_2$-normalize this representation before sending it to the softmax classification layer. All of these choices help cope with domain mismatch when the input is significantly different from the source domain.



\begin{figure*}[t]
  \centering
  \includegraphics[width=1.0\textwidth]{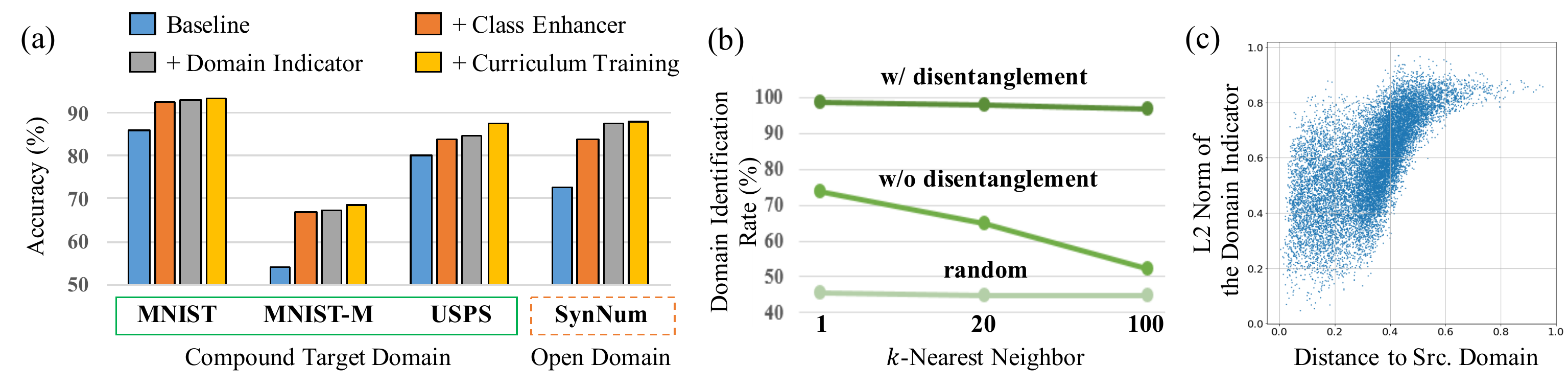}
  \vspace{-20pt}
  \caption{\textbf{Results of ablation studies about} (a) the memory-enhanced embeddings and curriculum domain adaptation, (b)  the domain-focused factors disentanglement, and (c) the memory-induced domain indicator vs.\  gaps to the source.}
  \label{fig:ablation}
\end{figure*}

\begin{table*}[t]
\footnotesize
\caption{\textbf{Performance on the C-Digits benchmark}. The methods in gray are especially designed for multi-target domain adaptation. $^{\dagger}$MTDA uses domain labels, while $^{\ddagger}$BTDA and DADA use the open domain images during training.}
\label{tab:benchmark_digits}
\parbox{0.3\linewidth}{
\centering
\includegraphics[width=0.27\textwidth]{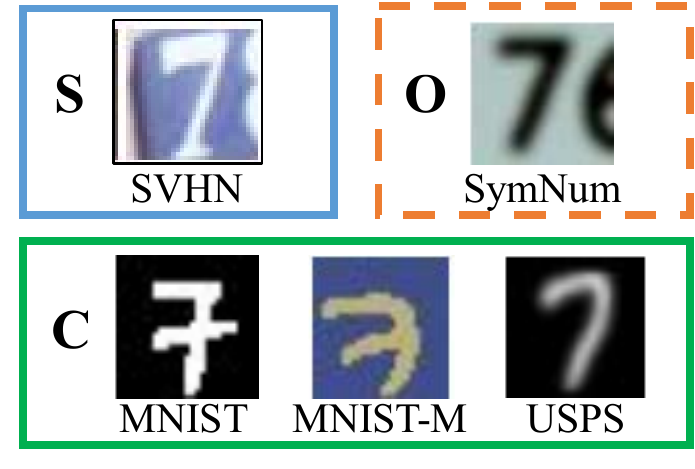}
}
\hfill
\parbox{0.7\linewidth}{
\centering
\begin{tabular}{l|ccc|c|cc}
\Xhline{1pt}
Src. Domain & \multicolumn{3}{c|}{Compound Domains (C)} & Open (O) & \multicolumn{2}{c}{Avg.} \\
\textbf{SVHN $\rightarrow$} & \textbf{MNIST} & \textbf{MNIST-M} & \textbf{USPS} & \textbf{SynNum} & \textbf{C} & \textbf{C+O} \\
\hline\hline
ADDA~\cite{tzeng2017adversarial} & 80.1$\pm$0.4 & 56.8$\pm$0.7 & 64.8$\pm$0.3 & 72.5$\pm$1.2 & 67.2$\pm$0.5 & 68.6$\pm$0.7 \\
JAN~\cite{long2017deep} & 65.1$\pm$0.1 & 43.0$\pm$0.1 & 63.5$\pm$0.2 & 85.6$\pm$0.0 & 57.2$\pm$0.1 & 64.3$\pm$0.1 \\
MCD~\cite{saito2018maximum} & 69.6$\pm$1.4 & 48.6$\pm$0.5 & 70.6$\pm$0.2 & \textbf{89.8$\pm$2.9} & 62.9$\pm$1.0 & 69.9$\pm$1.3 \\
\hline
\rowcolor{Gray}
MTDA$^{\dagger}$~\cite{gholami2018unsupervised} & 84.6$\pm$0.3 & 65.3$\pm$0.2 & 70.0$\pm$0.2 & - & 73.3$\pm$0.2 & - \\
\rowcolor{Gray}
BTDA$^{\ddagger}$~\cite{Chen_2019_CVPR} & 85.2$\pm$1.6 & \textbf{65.7$\pm$1.3} & 74.3$\pm$0.9 & 84.4$\pm$2.2 & 75.1$\pm$1.3 & 77.4$\pm$1.5 \\
\rowcolor{Gray}
DADA$^{\ddagger}$~\cite{peng2019domain} & - & - & - & - & - & 80.1$\pm$0.4 \\
\hline
Ours & \textbf{90.9$\pm$0.2} & \textbf{65.7$\pm$0.5} & \textbf{83.4$\pm$0.3} & 88.2$\pm$0.8 & \textbf{80.0$\pm$0.3} & \textbf{82.1$\pm$0.5} \\
\Xhline{1pt}
\end{tabular}
}
\end{table*}

\section{Experiments}

\noindent
\textbf{Datasets.}
To facilitate a comprehensive evaluation on various tasks (\ie, classification, segmentation, and navigation),  we carefully design four open compound domain adaptation (OCDA) benchmarks: C-Digits, C-Faces, C-Driving, and C-Mazes, respectively.
\begin{enumerate}[leftmargin=*]
\item \emph{C-Digits}: This benchmark aims to evaluate the classification adaptation ability under different appearances and backgrounds. It is built upon five classic digits datasets (SVHN~\cite{Yuval2011svhn}, MNIST~\cite{lecun1998gradient}, MNIST-M~\cite{ganin2015unsupervised}, USPS~\cite{Hull1994usps} and SynNum~\cite{ganin2015unsupervised}), where SVHN is used as the source domain, MNIST, MNIST-M, and USPS are mixed as the compound target domain, and SynNum is the open domain. We employ SWIT~\cite{swit} as an additional open domain for further analysis.
\item \emph{C-Faces}: This benchmark aims to evaluate the classification adaptation ability under different camera poses. It is built upon the Multi-PIE dataset~ \cite{Gross2008multipie}, where C05 (frontal view) is used as source domain, C08-C14 (left side view) are combined as the  compound target domain, and C19 (right side view) is kept out as the open domain.
\item \emph{C-Driving}: This benchmark aims to evaluate the segmentation adaptation ability from simulation to different real driving scenarios. The GTA-5~\cite{Richter_2016_ECCV} dataset is adopted as the source domain, while the BDD100K dataset~\cite{yu2018bdd100k} (with different scenarios including ``rainy'', ``snowy'', ``cloudy'', and ``overcast'') is taken for the compound and open domains.
\item \emph{C-Mazes}: This benchmark aims to evaluate the navigation adaptation ability under different environmental appearances. It is built upon the GridWorld environment~\cite{hu2018synthesized}, where mazes with different colors are used as the source and open domains. Since reinforcement learning often assumes no prior access to the environments, there are no compound target domains here.
\end{enumerate}

\noindent
\textbf{Network Architectures.}
To make a fair comparison with previous works~\cite{tzeng2017adversarial, gholami2018unsupervised, peng2019domain}, the modified LeNet-5~\cite{lecun1998gradient} and ResNet-18~\cite{he2016deep} are used as the backbone networks for C-Digits and C-Faces, respectively.
Following~\cite{tsai2018learning, zou2018unsupervised, pan2018two}, a  pre-trained VGG-16~\cite{Simonyan15} is the backbone network for C-Driving.
We additionally test our approach on reinforcement learning using ResNet-18 following~\cite{hu2018synthesized}.

\vspace{2pt}
\noindent
\textbf{Evaluation Metrics.}
The C-digits performance is measured by the digit classification accuracy, and the C-Faces performance is measured by the facial expression classification accuracy.
The C-Driving performance is measured by the standard mIOU, and the C-Mazes performance is measured by the average successful rate in 300 steps.
We evaluate the performance of each method with five runs and report both the mean and standard deviation. Moreover, we report both results of individual domains and the averaged results for a comprehensive analysis.

\vspace{2pt}
\noindent
\textbf{Comparison Methods.}
For classification tasks, we choose for comparison state-of-the-art methods in both conventional unsupervised domain adaptation (ADDA~\cite{tzeng2017adversarial}, JAN~\cite{long2017deep}, MCD~\cite{saito2018maximum}) and the recent multi-target domain adaptation methods (MTDA~\cite{gholami2018unsupervised}, BTDA~\cite{Chen_2019_CVPR}, DADA~\cite{peng2019domain}). 
Since MTDA~\cite{gholami2018unsupervised}, BTDA~\cite{Chen_2019_CVPR} and DADA~\cite{peng2019domain} are the most related to our work, we directly contrast our results to the numbers reported in their papers.
For the segmentation task, we compare with three state-of-the-art methods, AdaptSeg~\cite{tsai2018learning}, CBST~\cite{zou2018unsupervised}, IBN-Net~\cite{pan2018two} and PyCDA~\cite{lian2019constructing}.
For the reinforcement learning task, we benchmark with MTL, MLP~\cite{hu2018synthesized} and SynPo~\cite{hu2018synthesized}, a representative work for adaptation across environments.
We apply these methods to the same backbone networks as ours for a fair comparison.

\begin{table*}[t]
\footnotesize
\caption{\textbf{Performance on the C-Faces benchmark}. The methods in gray are especially designed for multi-target domain adaptation. $^{\dagger}$MTDA uses domain labels during training.}
\label{tab:benchmark_faces}
\vspace{-8pt}
\parbox{0.25\linewidth}{
\centering
\includegraphics[width=0.25\textwidth]{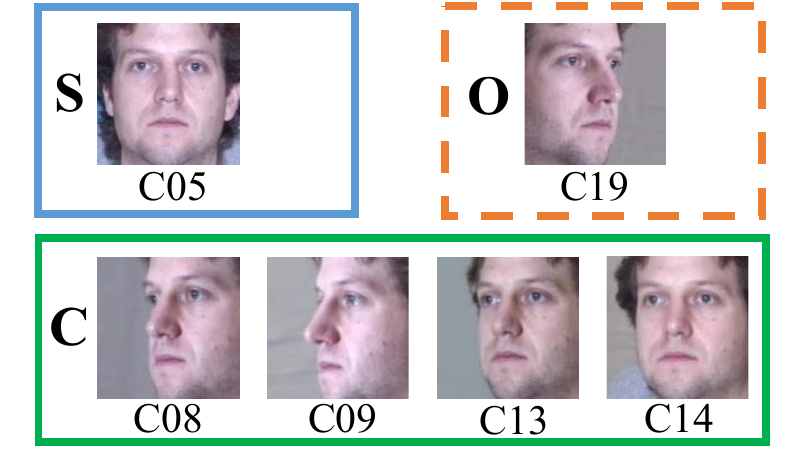}
}
\hfill
\parbox{0.78\linewidth}{
\centering
\begin{tabular}{l|cccc|c|cc}
\Xhline{1pt}
Src. Domain & \multicolumn{4}{c|}{Compound Domains (C)} & Open (O) & \multicolumn{2}{c}{Avg.} \\
\textbf{C05 $\rightarrow$} & \textbf{C08} & \textbf{C09} & \textbf{C13} & \textbf{C14} & \textbf{C19} & \textbf{C} & \textbf{C+O} \\
\hline\hline
ADDA~\cite{tzeng2017adversarial} & 46.9$\pm$0.2 & 36.4$\pm$0.5 & 39.1$\pm$0.3 & 65.4$\pm$0.4 & 71.8$\pm$0.8 & 47.0$\pm$0.4 & 51.9$\pm$0.4 \\
JAN~\cite{long2017deep} & 63.5$\pm$0.3 & 40.6$\pm$1.0 & 83.5$\pm$0.4 & \textbf{92.0$\pm$0.8} & 52.5$\pm$1.5 & 69.7$\pm$0.6 & 66.2$\pm$0.8 \\
MCD~\cite{saito2018maximum} & 50.4$\pm$0.5 & 45.8$\pm$0.2 & 77.8$\pm$0.1 & 88.0$\pm$0.1 & 60.4$\pm$0.9 & 65.7$\pm$0.2 & 64.6$\pm$0.4 \\
\hline
\rowcolor{Gray}
MTDA$^{\dagger}$~\cite{gholami2018unsupervised} & 49.0$\pm$0.2 & 48.2$\pm$0.1 & 53.1$\pm$0.2 & 84.3$\pm$0.1 & - & 58.7$\pm$0.2 & - \\
\hline
Ours & \textbf{73.3$\pm$0.2} & \textbf{55.1$\pm$0.4} & \textbf{84.1$\pm$0.1} & 88.9$\pm$0.3 & \textbf{72.7$\pm$0.6} & \textbf{75.4$\pm$0.3} & \textbf{74.8$\pm$0.3} \\
\Xhline{1pt}
\end{tabular}
}
\end{table*}

\begin{table*}[t]
\footnotesize
\caption{\textbf{Performance on the C-Driving ({left}) and C-Mazes benchmarks ({right})}. ``SynPo+Aug.'' indicates that we equip SynPo with proper color augmentation/randomization during training. Visual illustrations of both datasets are in Figure~\ref{fig:benchmark}.} 
\label{tab:benchmark_driving}
\vspace{-8pt}
\parbox{.5\linewidth}{
\centering
\begin{tabular}{l|ccc|c|cc}
\Xhline{1pt}
Source & \multicolumn{3}{c|}{Compound (C)} & Open (O) & \multicolumn{2}{c}{Avg.} \\
\textbf{GTA-5 $\rightarrow$} & \textbf{Rainy} & \textbf{Snowy} & \textbf{Cloudy} & \textbf{Overcast} & \textbf{C} & \textbf{C+O} \\
\hline\hline
Source Only & 16.2 & 18.0 & 20.9 & 21.2 & 18.9 & 19.1 \\ 
\hline
AdaptSeg~\cite{tsai2018learning} & 20.2 & 21.2 & 23.8 & 25.1 & 22.1 & 22.5 \\
CBST~\cite{zou2018unsupervised} & 21.3 & 20.6 & 23.9 & 24.7 & 22.2 & 22.6 \\
IBN-Net~\cite{pan2018two} & 20.6 & 21.9 & 26.1 & 25.5 & 22.8 & 23.5 \\
PyCDA~\cite{lian2019constructing} & 21.7 & 22.3 & 25.9 & 25.4 & 23.3 & 23.8 \\
\hline
Ours & \textbf{22.0} & \textbf{22.9} & \textbf{27.0} & \textbf{27.9} & \textbf{24.5} & \textbf{25.0} \\
\Xhline{1pt}
\end{tabular}
}
\hfill
\parbox{.45\linewidth}{
\centering
\begin{tabular}{l|cccc|c}
\Xhline{1pt}
Source & \multicolumn{4}{c|}{Open(O)} & Avg. \\
\textbf{M0 $\rightarrow$} & \textbf{M1} & \textbf{M2} & \textbf{M3} & \textbf{M4} & \textbf{O} \\
\hline\hline
Source Only & 0$\pm$0 & 0$\pm$0 & 0$\pm$0 & 0$\pm$0 & 0$\pm$0 \\
\hline
MTL & 0$\pm$0 & 30$\pm$5 & 75$\pm$0 & 65$\pm$5 & 42.5$\pm$2.5 \\
MLP~\cite{hu2018synthesized} & 5$\pm$5 & 45$\pm$10 & 75$\pm$5 & 80$\pm$10 & 51.2$\pm$7.5 \\
SynPo~\cite{hu2018synthesized} & 5$\pm$5 & 30$\pm$20 & 80$\pm$5 & 30$\pm$5 & 36.3$\pm$8.8 \\
SynPo+Aug. & 0$\pm$5 & 40$\pm$10 & 95$\pm$5 & 45$\pm$5 & 45.0$\pm$6.3 \\
\hline
Ours & \textbf{80$\pm$2.5} & \textbf{75$\pm$10} & \textbf{85$\pm$5} & \textbf{90$\pm$5} & \textbf{82.5$\pm$5.6} \\
\Xhline{1pt}
\end{tabular}
}
\end{table*}

\subsection{Ablation Study}
\label{sec:ablation}


\noindent
\textbf{Effectiveness of the Domain-Focused Factors Disentanglement.}
Here we verify that the domain-focused factors disentanglement helps discover the latent structures in the compound target domain.
It is probed by the domain identification rate within the $k$-nearest neighbors found by different encodings.
Figure~\ref{fig:ablation} (b) shows that features produced by our disentanglement  have a much higher identification rate ($\sim$95\%) than the counterparts without disentanglement ($\sim$65\%).

\vspace{2pt}
\noindent
\textbf{Effectiveness of the Curriculum Domain Adaptation.}
Figure~\ref{fig:ablation} (a) also reveals that, in the compound domain, the curriculum training contributes to the performance on USPS more than MNIST and MNITS-M.
On the other hand, we can observe from Figure~\ref{fig:manifold} and Table~\ref{tab:benchmark_digits} that USPS is the furthest target domain from the source domain SVHN. 
It implies that curriculum domain adaptation makes it easy to adapt to the distant target domains through an easy-to-hard adaptation schedule.

\vspace{2pt}
\noindent
\textbf{Effectiveness of Memory-Enhanced Representations.}
Recall that the memory-enhanced representations consist of two main components: the enhancer coming from the memory and the domain indicator.
From Figure~\ref{fig:ablation} (a), we observe that the class enhancer leads to large improvements on all target domains.
It is because the enhancer from the  memory transfers useful semantic concepts to the input of any domain.
Another observation is that the domain indicator is the most effective on the open domain (``SynNum''), because it helps dynamically calibrate the representations by leveraging domain relations (Figure~\ref{fig:ablation} (c)).

\subsection{Comparison Results}


\noindent
\textbf{C-Digits.}
Table~\ref{tab:benchmark_digits} shows the  comparison performances of different methods.
We have the following observations.
Firstly, ADDA~\cite{tzeng2017adversarial} and JAN~\cite{long2017deep} boost the performance on the compound domain by enforcing global distribution alignment.
However, they also sacrifice the performance on the open domain since there is no built-in mechanism for  handling any new domains, ``overfitting'' the model to the seen domains. 
Secondly, MCD~\cite{saito2018maximum} improves the results on the open domain, but its accuracy degrades on the compound target domain.
Maximizing the classifier discrepancy increases the robustness to the open domain; however, it also fails to capture the fine-grained latent structure in the compound target domain.
Lastly, compared to other multi-target domain adaptation methods (MTDA~\cite{gholami2018unsupervised} and DADA~\cite{peng2019domain}), our approach discovers domain structures and performs domain-aware knowledge transfer, achieving substantial advantages on all the test domains.

\vspace{2pt}
\noindent
\textbf{C-Faces.}
Similar observations can be made on the C-Faces benchmark as shown in Table~\ref{tab:benchmark_faces}.
Since face representations are inherently hierarchical, JAN~\cite{long2017deep} demonstrates competitive results on C14 due to its layer-wise transferring strategy.
Under the domain shift with different camera poses, our approach still consistently outperforms other alternatives for both the compound and open domains.

\begin{figure*}[t]
  \centering
  \includegraphics[width=1.0\textwidth]{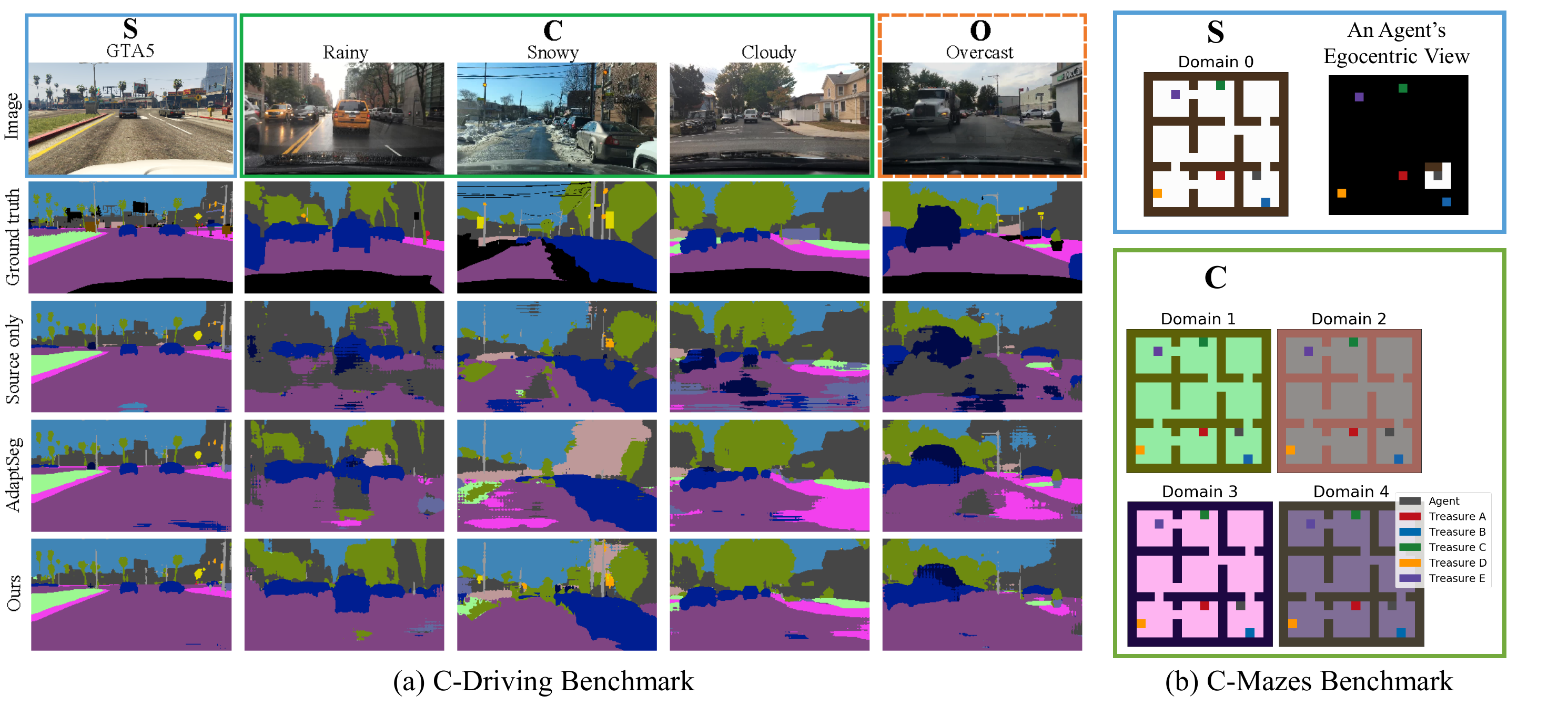}
  \vspace{-20pt}
  \caption{(a) \textbf{Qualitative results comparison} of semantic segmentation on the source domain (\textbf{S}), the compound target domain (\textbf{C}), and the open domain (\textbf{O}). (b) \textbf{Illustrations} of the 5 different domains in the C-Mazes benchmark. Our approach consistently outperforms existing domain adaptation methods across all compound and open target domains.}
  \label{fig:benchmark}
  \vspace{-10pt}
\end{figure*}

\vspace{2pt}
\noindent
\textbf{C-Driving.}
We compare with the state-of-the-art semantic segmentation adaptation methods such as AdaptSeg~\cite{tsai2018learning}, CBST~\cite{zou2018unsupervised}, and IBN-Net~\cite{pan2018two}.
All methods are tested under real-world driving scenarios in the BDD100K dataset~\cite{yu2018bdd100k}.
We can see that our approach has clear advantages on both the compound domain ($1.1\%$ gains) and the open domain ($2.4\%$ gains) as shown in Table~\ref{tab:benchmark_driving} ({left}).
We show detailed per-class accuracies in the appendices.
The qualitative comparisions are shown in Figure~\ref{fig:benchmark} (a).

\vspace{2pt}
\noindent
\textbf{C-Mazes.}
To directly compare with SynPo~\cite{hu2018synthesized}, we also evaluate on the GridWorld environments they provided.
The task in this benchmark is to learn navigation policies that can successfully collect all the treasures in the given mazes.
Existing reinforcement learning methods suffer from environmental changes, which we simulate as the appearances of the mazes here.
The final results are listed in Table~\ref{tab:benchmark_driving} ({right}).
Our approach transfers visual knowledge among navigation experiences and achieves more than $30\%$ improvements over the prior arts.

\subsection{Further Analysis}
\label{sec:analysis}

\begin{figure}
  \centering
  \includegraphics[width=0.48\textwidth]{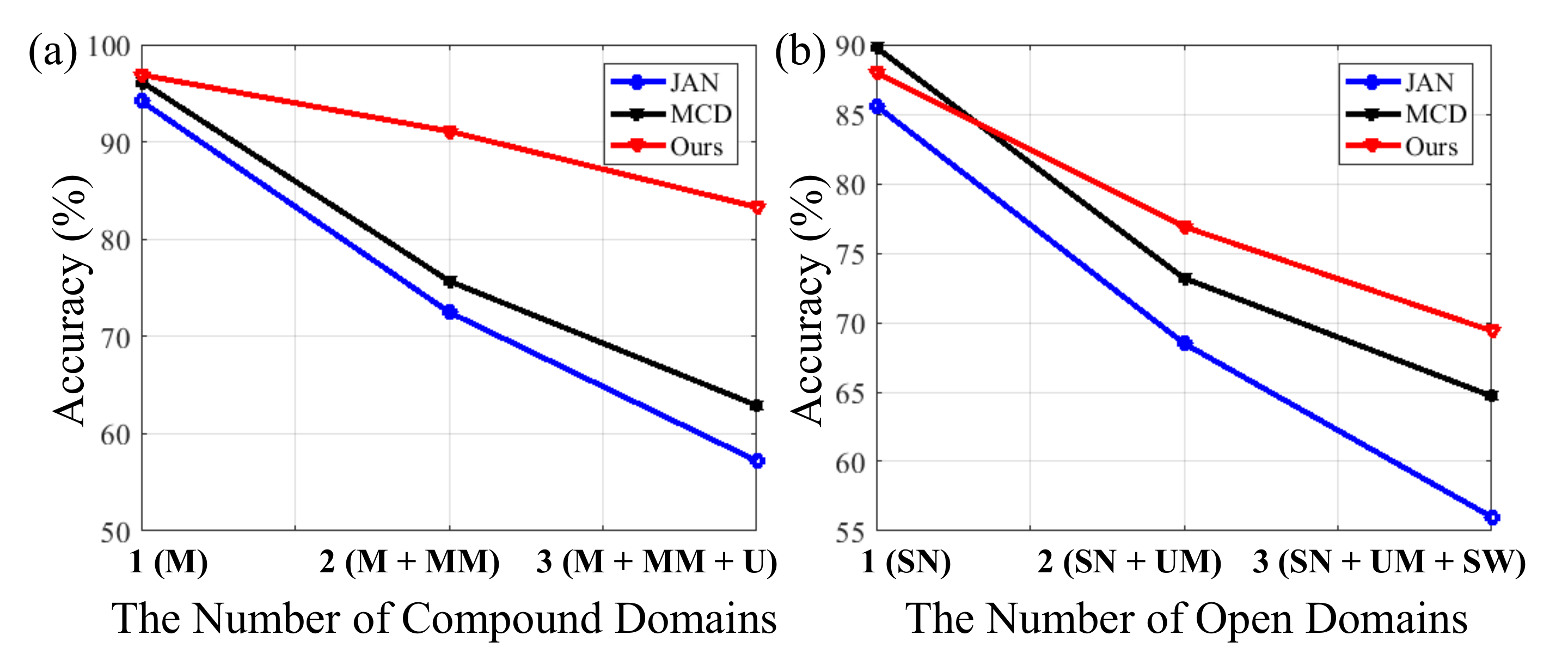}
  \vspace{-18pt}
  \caption{\textbf{Further analysis} on the (a) robustness to the complexity of the compound target domain and (b) robustness to the number of open domains. ``M'', ``MM'' and ``U'' stand for MNIST, MNIST-M, and USPS, respectively, while ``SN'', ``UM'' and ``SW'' stand for SynNum, USPS-M, and SWIT, respectively.}
  \label{fig:analysis}
\end{figure}

\noindent
\textbf{Robustness to the Complexity of the Compound Target Domain.}
We control the complexity of the compound target domain by varying the number of traditional target domains / datasets in it. 
Here we gradually increase constituting domains from a single target domain (\ie, MNIST) to two, and eventually three (\ie, MNIST + MNIST-M + USPS).
From Figure~\ref{fig:analysis} (a), we observe that as the number of datasets  increase, our approach only undergoes a moderate performance drop.
The learned curriculum enables gradual knowledge transfer that is capable of coping with complex  structures in the compound target domain.

\begin{figure}
  \centering
  \includegraphics[width=0.48\textwidth]{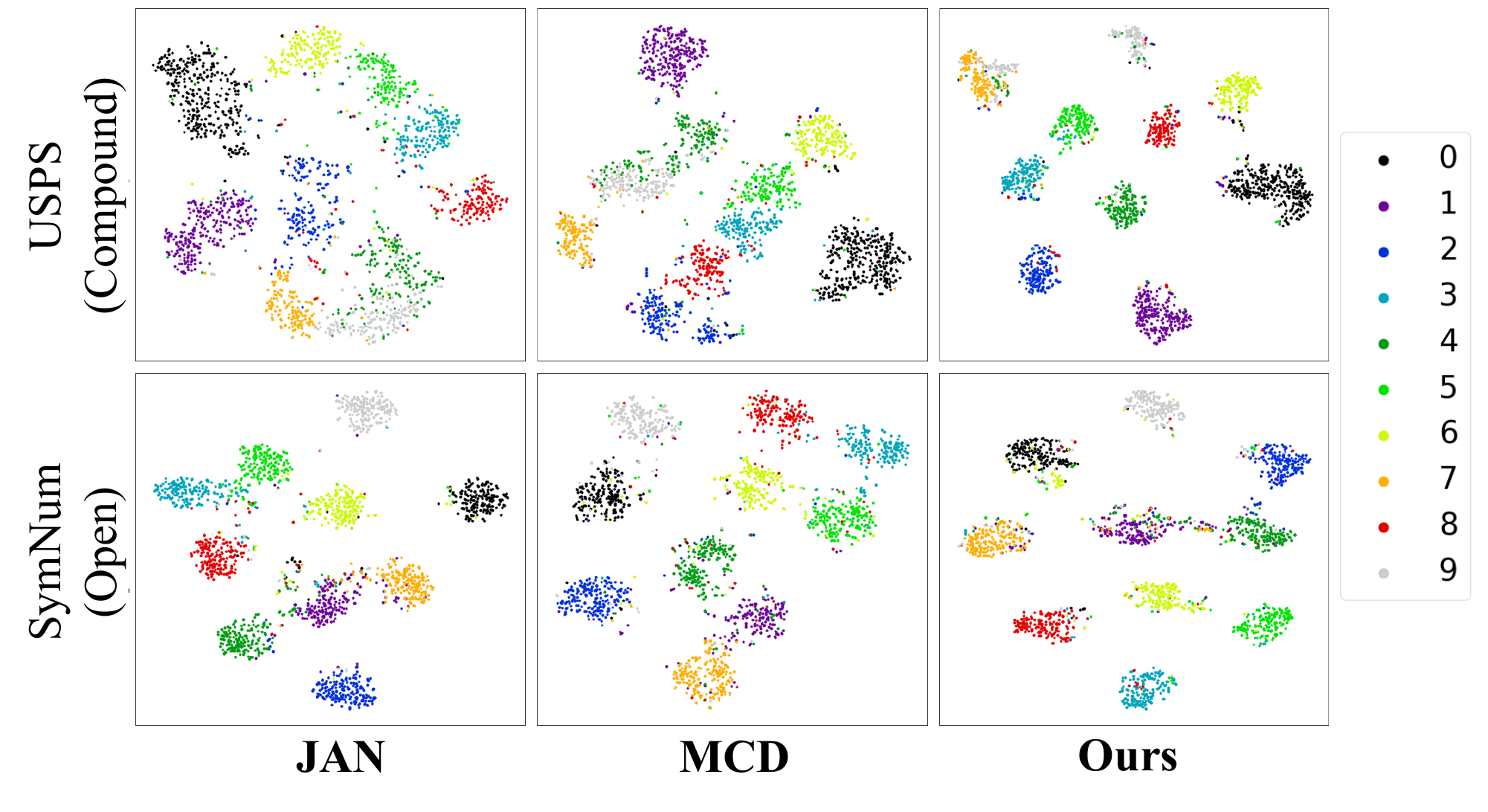}
  \vspace{-18pt}
  \caption{\textbf{t-SNE visualization} of the obtained embeddings. Compared to other methods, our approach is capable of producing class-discriminative features on both compound and open target domains.}
  \label{fig:tsne}
\end{figure}

\vspace{2pt}
\noindent
\textbf{Robustness to the Number of Open Domains.}
The performance change \wrt the number of open domains is demonstrated in Figure~\ref{fig:analysis} (b).
Here we include two new digits datasets, USPS-M (crafted in a similar way as MNIST-M) and SWIT~\cite{swit}, as the additional open domains.
Compared to JAN~\cite{long2017deep} and MCD~\cite{saito2018maximum}, our approach is more resilient to the various numbers of open domains.
The domain indicator module in our framework helps dynamically calibrate the embedding, thus enhancing the robustness to open domains.
Figure~\ref{fig:tsne} presents the t-SNE visualization comparison between the obtained embeddings of JAN~\cite{long2017deep}, MCD~\cite{saito2018maximum}, and our approach.

\section{Summary}

We formalize a more realistic topic called open compound domain adaptation for domain-robust learning.  We propose a novel model which includes a self-organizing curriculum domain adaptation to bootstrap generalization and a memory enhanced feature representation to build agility towards open domains.  We develop several benchmarks on classification, recognition, segmentation, and reinforcement learning and demonstrate the effectiveness of our model.

{\small
\noindent\textbf{Acknowledgements.}
This research was supported, in part, by the General Research Fund (GRF) of Hong Kong (No. 14236516 \& No. 14203518), Berkeley Deep Drive, DARPA, NSF 1835539, and US Government fund through Etegent Technologies on Low-Shot Detection in Remote Sensing Imagery.
}

{\small
\bibliographystyle{ieee_fullname}
\bibliography{reference}

\begin{thebibliography}{10}\itemsep=-1pt

\bibitem{swit}
Switzerland handwritten digits dataset.
\newblock \url{https://github.com/kensanata/numbers}.
\newblock Accessed: 2019-03-15.

\bibitem{Bobu2018adapting}
Andreea Bobu, Eric Tzeng, Judy Hoffman, and Trevor Darrell.
\newblock Adapting to continuously shifting domains.
\newblock {\em ICLR Workshop}, 2018.

\bibitem{cao2018partial}
Zhangjie Cao, Lijia Ma, Mingsheng Long, and Jianmin Wang.
\newblock Partial adversarial domain adaptation.
\newblock In {\em ECCV}, 2018.

\bibitem{chen2014semantic}
Liang-Chieh Chen, George Papandreou, Iasonas Kokkinos, Kevin Murphy, and Alan~L
  Yuille.
\newblock Semantic image segmentation with deep convolutional nets and fully
  connected crfs.
\newblock {\em ICLR}, 2015.

\bibitem{Chen_2019_CVPR}
Ziliang Chen, Jingyu Zhuang, Xiaodan Liang, and Liang Lin.
\newblock Blending-target domain adaptation by adversarial meta-adaptation
  networks.
\newblock In {\em CVPR}, 2019.

\bibitem{SynRealDataFog19}
Dengxin Dai, Christos Sakaridis, Simon Hecker, and Luc {Van Gool}.
\newblock Model adaptation with synthetic and real data for semantic dense
  foggy scene understanding.
\newblock {\em IJCV}, 2019.

\bibitem{ganin2015unsupervised}
Yaroslav Ganin and Victor Lempitsky.
\newblock Unsupervised domain adaptation by backpropagation.
\newblock In {\em ICML}, 2015.

\bibitem{ganin2016domain}
Yaroslav Ganin, Evgeniya Ustinova, Hana Ajakan, Pascal Germain, Hugo
  Larochelle, Fran{\c{c}}ois Laviolette, Mario Marchand, and Victor Lempitsky.
\newblock Domain-adversarial training of neural networks.
\newblock {\em JMLR}, 2016.

\bibitem{gholami2018unsupervised}
Behnam Gholami, Pritish Sahu, Ognjen Rudovic, Konstantinos Bousmalis, and
  Vladimir Pavlovic.
\newblock Unsupervised multi-target domain adaptation: An information theoretic
  approach.
\newblock {\em arXiv preprint arXiv:1810.11547}, 2018.

\bibitem{gidaris2018dynamic}
Spyros Gidaris and Nikos Komodakis.
\newblock Dynamic few-shot visual learning without forgetting.
\newblock In {\em CVPR}, 2018.

\bibitem{gong2013reshaping}
Boqing Gong, Kristen Grauman, and Fei Sha.
\newblock Reshaping visual datasets for domain adaptation.
\newblock In {\em NIPS}, 2013.

\bibitem{gong2012geodesic}
Boqing Gong, Yuan Shi, Fei Sha, and Kristen Grauman.
\newblock Geodesic flow kernel for unsupervised domain adaptation.
\newblock In {\em CVPR}, 2012.

\bibitem{gong2018dlow}
Rui Gong, Wen Li, Yuhua Chen, and Luc Van~Gool.
\newblock Dlow: Domain flow for adaptation and generalization.
\newblock In {\em CVPR}, 2019.

\bibitem{Gross2008multipie}
R. {Gross}, I. {Matthews}, J. {Cohn}, T. {Kanade}, and S. {Baker}.
\newblock Multi-pie.
\newblock In {\em 2008 8th IEEE International Conference on Automatic Face
  Gesture Recognition}, 2008.

\bibitem{he2016deep}
Kaiming He, Xiangyu Zhang, Shaoqing Ren, and Jian Sun.
\newblock Deep residual learning for image recognition.
\newblock In {\em CVPR}, 2016.

\bibitem{hoffman2012discovering}
Judy Hoffman, Brian Kulis, Trevor Darrell, and Kate Saenko.
\newblock Discovering latent domains for multisource domain adaptation.
\newblock In {\em ECCV}, 2012.

\bibitem{hoffman2017cycada}
Judy Hoffman, Eric Tzeng, Taesung Park, Jun-Yan Zhu, Phillip Isola, Kate
  Saenko, Alexei~A Efros, and Trevor Darrell.
\newblock Cycada: Cycle-consistent adversarial domain adaptation.
\newblock In {\em ICML}, 2018.

\bibitem{hu2018synthesized}
Hexiang Hu, Liyu Chen, Boqing Gong, and Fei Sha.
\newblock Synthesized policies for transfer and adaptation across tasks and
  environments.
\newblock In {\em NIPS}, 2018.

\bibitem{Hull1994usps}
J.~J. {Hull}.
\newblock A database for handwritten text recognition research.
\newblock {\em TPAMI}, 1994.

\bibitem{kingma2014adam}
Diederik~P Kingma and Jimmy Ba.
\newblock Adam: A method for stochastic optimization.
\newblock {\em ICLR}, 2015.

\bibitem{lecun1998gradient}
Yann LeCun, L{\'e}on Bottou, Yoshua Bengio, Patrick Haffner, et~al.
\newblock Gradient-based learning applied to document recognition.
\newblock {\em Proceedings of the IEEE}, 1998.

\bibitem{li2018learning}
Da Li, Yongxin Yang, Yi-Zhe Song, and Timothy~M Hospedales.
\newblock Learning to generalize: Meta-learning for domain generalization.
\newblock In {\em AAAI}, 2018.

\bibitem{li2018domain}
Haoliang Li, Sinno Jialin~Pan, Shiqi Wang, and Alex~C Kot.
\newblock Domain generalization with adversarial feature learning.
\newblock In {\em CVPR}, 2018.

\bibitem{li2017domain}
Wen Li, Zheng Xu, Dong Xu, Dengxin Dai, and Luc Van~Gool.
\newblock Domain generalization and adaptation using low rank exemplar svms.
\newblock {\em TPAMI}, 2017.

\bibitem{li2017not}
Xiaoxiao Li, Ziwei Liu, Ping Luo, Chen Change~Loy, and Xiaoou Tang.
\newblock Not all pixels are equal: Difficulty-aware semantic segmentation via
  deep layer cascade.
\newblock In {\em CVPR}, 2017.

\bibitem{lian2019constructing}
Qing Lian, Fengmao Lv, Lixin Duan, and Boqing Gong.
\newblock Constructing self-motivated pyramid curriculums for cross-domain
  semantic segmentation: A non-adversarial approach.
\newblock In {\em ICCV}, 2019.

\bibitem{liu2017sphereface}
Weiyang Liu, Yandong Wen, Zhiding Yu, Ming Li, Bhiksha Raj, and Le Song.
\newblock Sphereface: Deep hypersphere embedding for face recognition.
\newblock In {\em CVPR}, 2017.

\bibitem{liu2019large}
Ziwei Liu, Zhongqi Miao, Xiaohang Zhan, Jiayun Wang, Boqing Gong, and Stella~X
  Yu.
\newblock Large-scale long-tailed recognition in an open world.
\newblock In {\em CVPR}, 2019.

\bibitem{liu2016fashion}
Ziwei Liu, Sijie Yan, Ping Luo, Xiaogang Wang, and Xiaoou Tang.
\newblock Fashion landmark detection in the wild.
\newblock In {\em ECCV}, 2016.

\bibitem{long2017deep}
Mingsheng Long, Han Zhu, Jianmin Wang, and Michael~I Jordan.
\newblock Deep transfer learning with joint adaptation networks.
\newblock In {\em ICML}, 2017.

\bibitem{Mancini_2019_CVPR}
Massimiliano Mancini, Samuel~Rota Bulo, Barbara Caputo, and Elisa Ricci.
\newblock Adagraph: Unifying predictive and continuous domain adaptation
  through graphs.
\newblock In {\em CVPR}, 2019.

\bibitem{mancini2019inferring}
Massimiliano Mancini, Lorenzo Porzi, Samuel~Rota Bulo, Barbara Caputo, and
  Elisa Ricci.
\newblock Inferring latent domains for unsupervised deep domain adaptation.
\newblock {\em TPAMI}, 2019.

\bibitem{Yuval2011svhn}
Yuval Netzer, Tao Wang, Adam Coates, Alessandro Bissacco, Bo Wu, and Andrew
  Y~Ng.
\newblock Reading digits in natural images with unsupervised feature learning.
\newblock {\em NIPS}, 2011.

\bibitem{pan2010domain}
Sinno~Jialin Pan, Ivor~W Tsang, James~T Kwok, and Qiang Yang.
\newblock Domain adaptation via transfer component analysis.
\newblock {\em IEEE Transactions on Neural Networks}, 2010.

\bibitem{pan2018two}
Xingang Pan, Ping Luo, Jianping Shi, and Xiaoou Tang.
\newblock Two at once: Enhancing learning and generalization capacities via
  ibn-net.
\newblock In {\em ECCV}, 2018.

\bibitem{Pan_2019_CVPR}
Yingwei Pan, Ting Yao, Yehao Li, Yu Wang, Chong-Wah Ngo, and Tao Mei.
\newblock Transferrable prototypical networks for unsupervised domain
  adaptation.
\newblock In {\em CVPR}, 2019.

\bibitem{panareda2017open}
Pau Panareda~Busto and Juergen Gall.
\newblock Open set domain adaptation.
\newblock In {\em ICCV}, 2017.

\bibitem{peng2019moment}
Xingchao Peng, Qinxun Bai, Xide Xia, Zijun Huang, Kate Saenko, and Bo Wang.
\newblock Moment matching for multi-source domain adaptation.
\newblock In {\em ICCV}, 2019.

\bibitem{peng2019domain}
Xingchao Peng, Zijun Huang, Ximeng Sun, and Kate Saenko.
\newblock Domain agnostic learning with disentangled representations.
\newblock In {\em ICML}, 2019.

\bibitem{Richter_2016_ECCV}
Stephan~R. Richter, Vibhav Vineet, Stefan Roth, and Vladlen Koltun.
\newblock Playing for data: {G}round truth from computer games.
\newblock In {\em ECCV}, 2016.

\bibitem{saenko2010adapting}
Kate Saenko, Brian Kulis, Mario Fritz, and Trevor Darrell.
\newblock Adapting visual category models to new domains.
\newblock In {\em ECCV}, 2010.

\bibitem{saito2018maximum}
Kuniaki Saito, Kohei Watanabe, Yoshitaka Ushiku, and Tatsuya Harada.
\newblock Maximum classifier discrepancy for unsupervised domain adaptation.
\newblock In {\em CVPR}, 2018.

\bibitem{saito2018open}
Kuniaki Saito, Shohei Yamamoto, Yoshitaka Ushiku, and Tatsuya Harada.
\newblock Open set domain adaptation by backpropagation.
\newblock In {\em ECCV}, 2018.

\bibitem{Simonyan15}
K. Simonyan and A. Zisserman.
\newblock Very deep convolutional networks for large-scale image recognition.
\newblock In {\em ICLR}, 2015.

\bibitem{snell2017prototypical}
Jake Snell, Kevin Swersky, and Richard Zemel.
\newblock Prototypical networks for few-shot learning.
\newblock In {\em NIPS}, 2017.

\bibitem{torralba2011}
Antonio Torralba and Alexei~A Efros.
\newblock Unbiased look at dataset bias.
\newblock In {\em CVPR}, 2011.

\bibitem{tsai2018learning}
Yi-Hsuan Tsai, Wei-Chih Hung, Samuel Schulter, Kihyuk Sohn, Ming-Hsuan Yang,
  and Manmohan Chandraker.
\newblock Learning to adapt structured output space for semantic segmentation.
\newblock In {\em CVPR}, 2018.

\bibitem{tzeng2017adversarial}
Eric Tzeng, Judy Hoffman, Kate Saenko, and Trevor Darrell.
\newblock Adversarial discriminative domain adaptation.
\newblock In {\em CVPR}, 2017.

\bibitem{venkateswara2017deep}
Hemanth Venkateswara, Jose Eusebio, Shayok Chakraborty, and Sethuraman
  Panchanathan.
\newblock Deep hashing network for unsupervised domain adaptation.
\newblock In {\em CVPR}, 2017.

\bibitem{wu2019ace}
Zuxuan Wu, Xin Wang, Joseph~E Gonzalez, Tom Goldstein, and Larry~S Davis.
\newblock Ace: Adapting to changing environments for semantic segmentation.
\newblock In {\em ICCV}, 2019.

\bibitem{xiong2014latent}
Caiming Xiong, Scott McCloskey, Shao-Hang Hsieh, and Jason~J Corso.
\newblock Latent domains modeling for visual domain adaptation.
\newblock In {\em AAAI}, 2014.

\bibitem{xu2014exploiting}
Zheng Xu, Wen Li, Li Niu, and Dong Xu.
\newblock Exploiting low-rank structure from latent domains for domain
  generalization.
\newblock In {\em ECCV}, 2014.

\bibitem{yu2018bdd100k}
Fisher Yu, Wenqi Xian, Yingying Chen, Fangchen Liu, Mike Liao, Vashisht
  Madhavan, and Trevor Darrell.
\newblock Bdd100k: A diverse driving video database with scalable annotation
  tooling.
\newblock {\em arXiv preprint arXiv:1805.04687}, 2018.

\bibitem{yu2018multi}
Huanhuan Yu, Menglei Hu, and Songcan Chen.
\newblock Multi-target unsupervised domain adaptation without exactly shared
  categories.
\newblock {\em arXiv preprint arXiv:1809.00852}, 2018.

\bibitem{zhang2018importance}
Jing Zhang, Zewei Ding, Wanqing Li, and Philip Ogunbona.
\newblock Importance weighted adversarial nets for partial domain adaptation.
\newblock In {\em CVPR}, 2018.

\bibitem{zhang2019curriculum}
Yang Zhang, Philip David, Hassan Foroosh, and Boqing Gong.
\newblock A curriculum domain adaptation approach to the semantic segmentation
  of urban scenes.
\newblock {\em TPAMI}, 2019.

\bibitem{zou2019confidence}
Yang Zou, Zhiding Yu, Xiaofeng Liu, BVK Kumar, and Jinsong Wang.
\newblock Confidence regularized self-training.
\newblock In {\em ICCV}, 2019.

\bibitem{zou2018unsupervised}
Yang Zou, Zhiding Yu, BVK Vijaya~Kumar, and Jinsong Wang.
\newblock Unsupervised domain adaptation for semantic segmentation via
  class-balanced self-training.
\newblock In {\em ECCV}, 2018.

\end{thebibliography}
}

\clearpage

\appendix
\addcontentsline{toc}{section}{Appendices}
\section*{Appendices}

\newcommand{\R}{\rotatebox[origin=c]{90}}

In this supplementary material, we provide details omitted in the main text including:
\begin{itemize}[leftmargin=*]
\item Section~\ref{sec:relation}: relation to other DA problems (Sec. 2 ``Related Works'' of the main paper.)
\item Section~\ref{sec:methodology}: more methodology details (Sec. 3 ``Our Approach'' of the main paper.)
\item Section~\ref{sec:experimental}: detailed experimental setup (Sec. 4 ``Experiments'' of the main paper.)
\item Section~\ref{sec:results}: additional comparison results (Sec. 4.2 ``Comparison Results'' of the main paper.)
\item Section~\ref{sec:visualization}: additional visualization of our approach (Sec. 4.3 ``Further Analysis'' of the main paper.)
\end{itemize}

\section{Relation to Other DA Problems}
\label{sec:relation}

Human vision system shows remarkable generalization ability to see clear across many different domains, such as in foggy and rainy days. Computer vision system, on the other hand, has long been haunted by this domain shift issue. Several sub-fields in domain adaptation have been studies to mitigate this challenge. 


\noindent
\textbf{Open/Partial Set Domain Adaptation.}
Another route of research aims to tackle the category sharing/unsharing issues between source and target domain, namely open set~\cite{panareda2017open, saito2018open} and partial set~\cite{zhang2018importance, cao2018partial} domain adaptation. They assume that the target domain contains either (1) new categories that don't appear in source domain; or (2) only a subset of categories that appear in source domain. Both settings concern the ``openness'' of categories. Instead, in this work we investigate the ``openness'' of domains, \ie we assume there are unknown domains existing that are absent in the training phase.

\section{More Methodology Details}
\label{sec:methodology}


\noindent
\textbf{Notation Summary.}
We summarize the notations used in the paper in Table~\ref{tab:notation}.

\begin{table}[ht]
    \caption{\textbf{Summary} of notations.}
    \label{tab:notation}
    \small
    \centering
    \begin{tabular}{l|c}
    \Xhline{1pt}
    ~{\bf Notation}~ & ~{\bf Meaning}~  \\ \hline \hline
    $x$ & input image \\ \hline
    $y$ & category label \\ \hline
    $z_{random}$ & random category label \\ \hline
    $E_{class}(\cdot)$ & class encoder \\ \hline
    $\Phi(\cdot)$ & class classifier \\ \hline
    $E_{domain}(\cdot)$ & domain encoder \\ \hline
    $Decoder(\cdot)$ & class decoder \\ \hline
    $D(\cdot)$ & class classifier after domain encoder\\ \hline
    $v_{direct}$ & direct feature \\ \hline
    $M$ & visual memory \\ \hline
    $v_{enhance}$ & class enhancer \\ \hline
    $e_{domain}$ & domain indicator \\ \hline
    $T(\cdot)$ & network that generates $e_{domain}$ \\ \hline
    $v_{transfer}$ & source-enhanced representation\\ \hline
    \Xhline{1pt}
    \end{tabular}
    \vspace{6pt}
\end{table}

\noindent
\textbf{Details of Class and Domain Manifold Disentanglement.}
The detailed class and domain manifold disentanglement algorithm is shown in Algorithm~\ref{alg:disentangle}.

\begin{algorithm}[ht]
    \caption{Disentangling training. $E_{class}(\cdot)$ and $\Phi$ has been trained using source-domain data, $Deccoder(\cdot)$: the decoder, $C$: number of classes, $\gamma$: a constant.}
    \label{alg:disentangle}
    \begin{algorithmic}
    \INPUT{$E_{class}(\cdot)$, $E_{domain}(\cdot)$, $\Phi(\cdot)$, $D(\cdot)$, $Decoder(\cdot)$, $C$, $\gamma$}
    \For{k iterations}
        \State Sample mini-batch $x$.
        \State Compute pseudo label $y_{pseudo} \gets \Phi\left(E_{class}\left(x\right)\right)$.
        \State Update the discriminator $D$ with: $\nabla_{\theta_{D}} \sum_j y_{pseudo}^j \log\left(D\left(E_{domain}\left(x^j\right)\right)\right)$.
        \State Prepare random label $z_{random} \sim uniform\{0,1,...,C-1\}$, and convert it to one-hot vector $y_{random}$.
        \State Compute adversarial loss: $L_{adv} \gets \sum_j y_{random}^j\log\left(D\left(E_{domain}(x^j)\right)\right)$.
        \State Compute reconstruction loss: $L_{rec} \gets \sum_j \lvert Decoder\left(E_{class}\left(x^j\right), E_{domain}\left(x^j\right)\right) - x^j\lvert$.
        \State Update the domain encoder $E_{domain}$ with: $\nabla_{\theta_{E_{domain}}} \left(L_{adv} + \gamma L_{rec}\right)$.
    \EndFor
    \end{algorithmic}
\end{algorithm}

\noindent
\textbf{Time Complexity.}
Our approach introduces negligible computational overhead (1.3\%) to the standard deep networks, such as VGG~\cite{Simonyan15} and ResNet~\cite{he2016deep}, since only a lightweight memory module is inserted during inference.

\begin{figure}[h]
  \centering
  \includegraphics[width=0.49\textwidth]{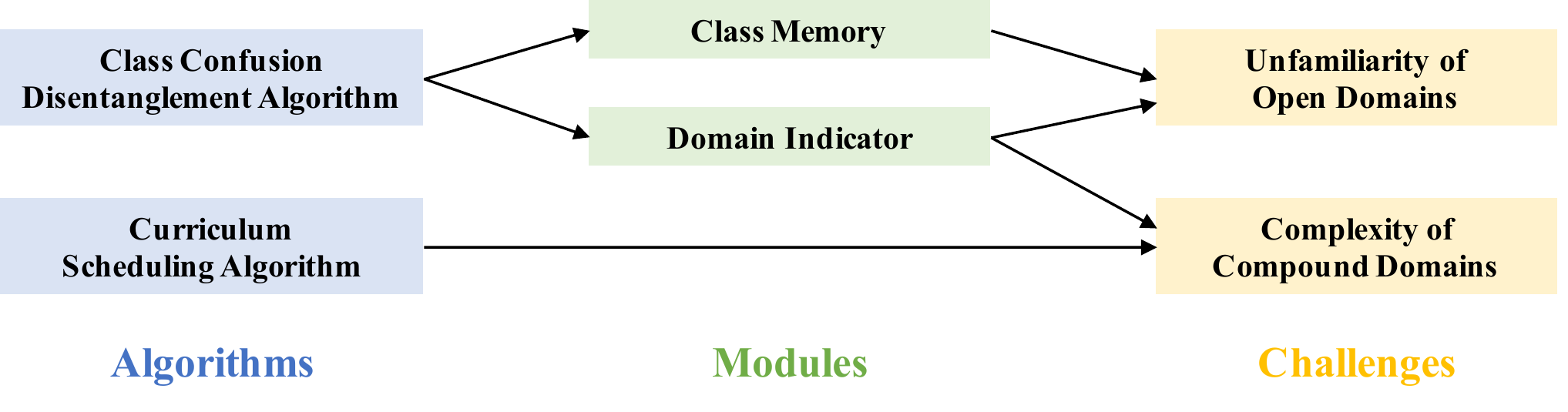}
  \caption{\textbf{Methodology highlight} of our entire neural architecture.}
  \label{fig:framework}
\end{figure}

\noindent
\textbf{Methodology Highlight.}
Our main methodology contribution is the entire neural architecture that can address the complexity of compound domains during training and handle the unseen domains during testing, as depicted in Figure~\ref{fig:framework}.

\noindent
\textbf{Methodology Comparisons.}
Table~\ref{tab:diff} summarizes the key methodological differences between MTDA~\cite{gholami2018unsupervised}, BTDA~\cite{Chen_2019_CVPR}, DADA~\cite{peng2019domain} and our OCDA.

\begin{table}[h]
    \scriptsize
    \centering
    \caption{\textbf{Methodology comparisons} between MTDA, BTDA, DADA and our OCDA.}
    \label{tab:diff}
    \begin{tabular}{l|c|c|c|c}
    \Xhline{1pt}
     & {\bf MTDA}~\cite{gholami2018unsupervised} & {\bf BTDA}~\cite{Chen_2019_CVPR} & {\bf DADA}~\cite{peng2019domain} & {\bf OCDA} \\ \hline
    \tabincell{c}{Feature\\Disentangle} & \tabincell{c}{entropy\\minimization} & $\times$ & \tabincell{c}{mutual\\information} & \tabincell{c}{class\\confusion} \\ \hline
    \tabincell{c}{Domain\\Invariance} & adversarial & adversarial & adversarial & \tabincell{c}{memory+\\adversarial} \\ \hline
    \tabincell{c}{Latent Domain\\Discovery} & $\times$ & clustering & $\times$ & \tabincell{c}{domain\\indicator} \\ \hline
    \tabincell{c}{Latent Domain\\Ranking} & $\times$ & $\times$ & $\times$ & curriculum \\ \hline
    \Xhline{1pt}
    \end{tabular}
\end{table}

\begin{table*}[h]
\caption{\textbf{Statistics of BDD100K dataset} in our Open Compound Domain Adaptation (OCDA) setting.}
\label{tab:dataset_driving}
\centering
\begin{tabular}{l|ccc|c|c}
\Xhline{1pt}
 & \multicolumn{3}{c|}{Compound (C)} & Open &  \\
{\textbf{Domains}} & {\textbf{Rainy}} & {\textbf{Snowy}} & {\textbf{Cloudy}} & {\textbf{Overcast}} & {\textbf{Total}} \\
\hline\hline
training set w/o label & 4855 & 5307 & 4535 & 8143 & 22840 \\
validation set w/ label & 215 & 242 & 346 & 627 & 1430 \\
\Xhline{1pt}
\end{tabular}
\end{table*}

\begin{figure*}[ht]
  \centering
  \includegraphics[width=1.0\textwidth]{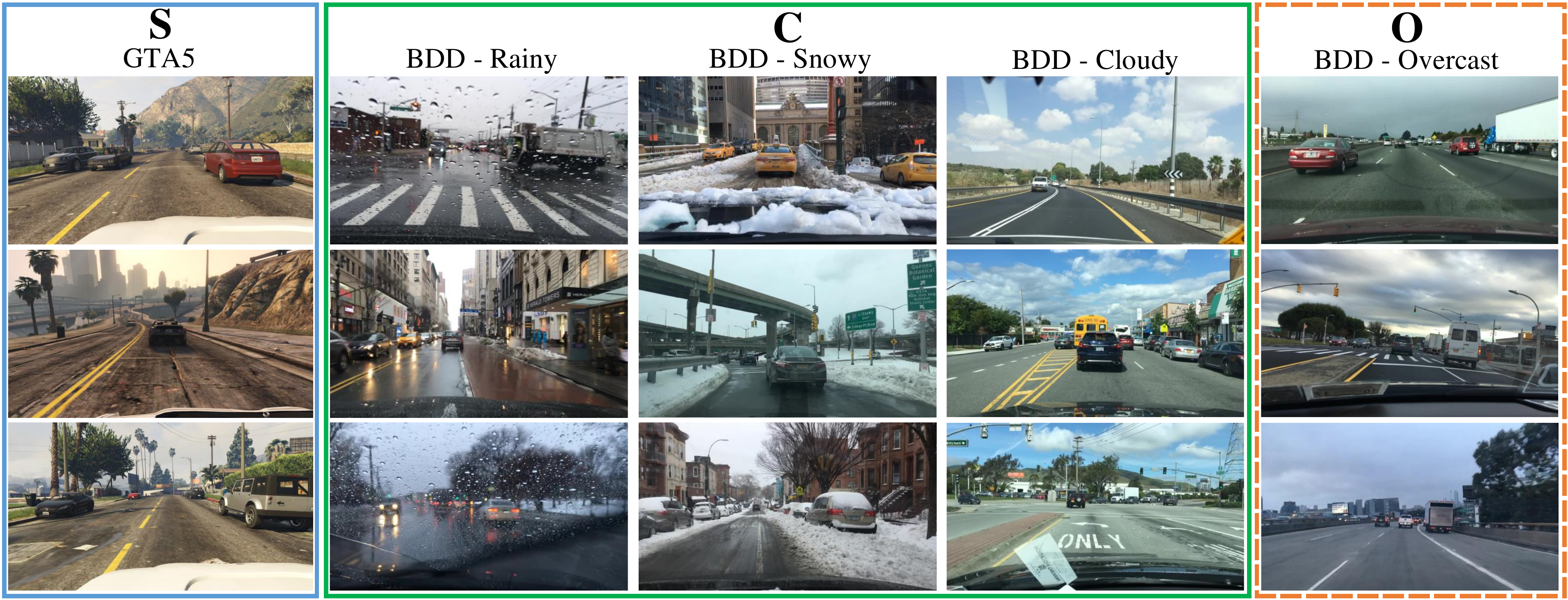}
  \caption{\textbf{Examples of the C-Driving benchmark} (GTA5 and BDD100K datasets). }
  \label{fig:dataset_drive}
\end{figure*}

\section{Experimental Setup}
\label{sec:experimental}

\subsection{OCDA Datasets and Benchmarks}

\noindent
\textbf{C-Digits.} 
The C-Digits dataset is consist of 5 digit datasets: SVHN \cite{Yuval2011svhn}, MNIST \cite{lecun1998gradient}, MNIST-M \cite{ganin2015unsupervised}, USPS \cite{Hull1994usps}, and SynNum \cite{ganin2015unsupervised}. SVHN is a dataset of street view housing numbers. MNIST and USPS are two datasets of hand-written numbers. MNIST-M and SynNum are two datasets of synthetic numbers. We choose SVHN as the source domain, MNIST, MNIST-M, and USPS as target domains that can be accessed during training, and SynNum as the open domain, which is only accessible during testing. Images from all domains are scaled to $32 \times 32$ and converted to RGB format. We follow the origin training/testing split of each dataset. The SVHN dataset is balanced across the classes, following \cite{hoffman2017cycada}. Besides, no further pre-processing is applied to the data.

\noindent
\textbf{C-Faces.}
We choose images of 6 different view angles from the We choose images of 6 different view angles from the Multi-PIE dataset \cite{Gross2008multipie}: C051, C080, C090, C130, C140, and C190. 6 facial expressions are treated as classification categories: neutral, smile, surprise, squint, disgust, and scream. We use images of view angle C051 as the source domain in our experiment, C080, C090, C130, and C140 as target domains that can be accessed during training, and C190 as the open domain. All the images are aligned according to the face landmarks and cropped around the facial areas into $224 \times 224$ images. 
\begin{figure*}[ht]
  \centering
  \includegraphics[width=1.0\textwidth]{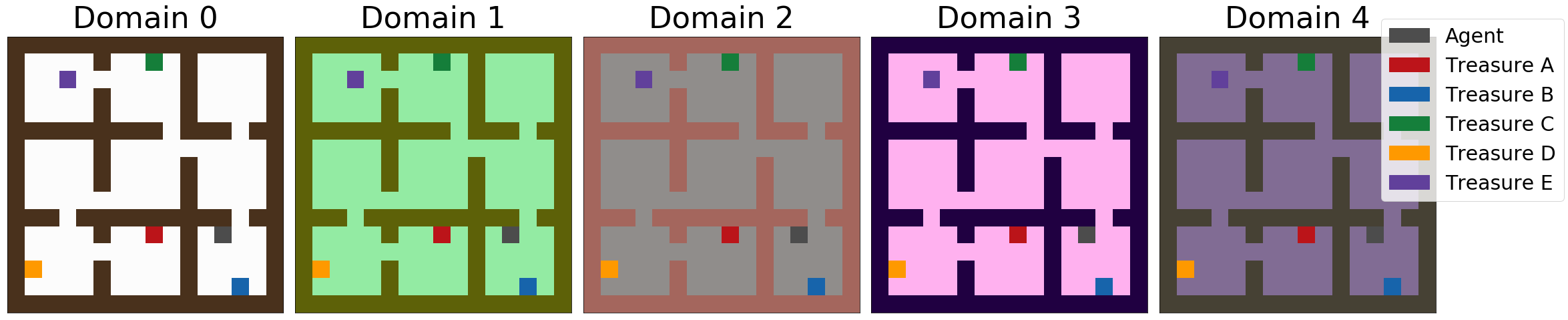}
  \caption{\textbf{Illustrations} of the five different domains in the C-Mazes benchmark.}
  \label{fig:C-Mazes}
\end{figure*}

\noindent
\textbf{C-Driving.}
For semantic segmentation on driving scenarios, we adopt the \textbf{GTA5}~\cite{Richter_2016_ECCV} dataset as the source domain, and the \textbf{BDD100K}~\cite{yu2018bdd100k} dataset as the Compound and open domains. 
Examples of these datasets are illustrated in Figure~\ref{fig:dataset_drive}.

GTA5 is a virtual street view dataset generated from Grand Theft Auto V (GTA5). 
It has 24966 images of resolution $1914 \times 1052$.
The domain categories and statistics of the BDD100K dataset are provided in Table~\ref{tab:dataset_driving}.
We use the \textit{Rainy}, \textit{Snowy}, \textit{Cloudy}, and \textit{Overcast} domains in the training set of BDD100K in our experiments.
The \textit{Overcast} domain is used as the open domain while the rest are used as the Compound domains.
Other domains of the original BDD100K dataset like \textit{Clear} and \textit{Foggy} are not used.
Among these data, a small fraction with annotations is used as the validation set, while the rest are used as the training set for adaptation.
The results in our experiments are reported on the validation set.

\noindent
\textbf{C-Mazes.} 
This is a reinforcement learning dataset which is consist of mazes with different colors. The mazes are generated following \cite{hu2018synthesized}, where agent is asked to collect treasure spots from the mazes by different orders. For simplicity, we only use one scene (\ie maze with one topology) and two tasks (\ie two different treasure collection orders) for the experiments. We randomly generated 5 combinations of colors to construct the domains (which are illustrated in Figure~\ref{fig:C-Mazes}). The colors of agent and five treasures are the same across the domain. We use \textbf{Domain 0} as the source domain, and the rest of the 4 domains are target domains.

\subsection{Training Details}

\noindent
\textbf{C-Digits.}
The backbone model for this experiment is a LeNet-5~\cite{lecun1998gradient}, following the setups in \cite{hoffman2017cycada}. There are two stages in the training process. (1) In the first stage, we train the network with discriminative centroid loss for 100 epochs as a warm start for the memory-enhanced deep neural network. (2) In the second stage, we further fine-tune the networks with curriculum sampling and memory modules on top of the backbone network, where a domain adversarial loss~\cite{tzeng2017adversarial} is incorporated and the weights learned in the first stage are copied to two identical but independent networks (source network and target network). Only weights of the target network are updated during this stage. Centroids of each class (\ie constituting elements in the class memory) are calculated in the beginning of this stage, and the classifiers are reinitialized. The model is trained without the discriminative centroid loss in stage 2. Some major hyper-parameters can be found in Table~\ref{tab:parameters}.

\noindent
\textbf{C-Faces.}
Experiments on the C-Faces dataset are similar to experiments on the C-Digits dataset. However, the backbone model is ResNet18~\cite{he2016deep} with random initialization, instead of a LeNet-5. Some major hyper-parameters can also be found in Table~\ref{tab:parameters}.

\begin{table}[h]
    \vspace{6pt}
    \caption{\textbf{The major hyper-parameters} used in our experiments. ``LR.'' stands for learning rate.}
    \label{tab:parameters}
    \footnotesize
    \centering
    \begin{tabular}{l|c|c|c}
    \Xhline{1pt}
    ~{\bf Dataset}~ & ~{\bf Initial LR.}~ & ~{\bf Epoch}~ & ~{\bf betas for ADAM}~ \\ \hline \hline
    C-Digits (stage 1) & 1e-4  & 100 & (0.9, 0.999) \\ \hline
    C-Digits (stage 2) & 1e-5  & 200 & (0.9, 0.999) \\ \hline
    C-Faces (stage 1) & 1e-4  & 100 & (0.9, 0.999) \\ \hline
    C-Faces (stage 2) & 1e-5  & 200 & (0.9, 0.999) \\ \hline
    \Xhline{1pt}
    \end{tabular}
    \vspace{6pt}
\end{table}

\noindent
\textbf{C-Driving.}
Our implementation mainly follows \cite{tsai2018learning}.
We use DeepLab-VGG16~\cite{chen2014semantic} model with synchronized batch normalization and the batch size is set to 8.
The initial learning rate is 0.01 and is decreased using the "poly" policy with 0.9 power.
The maximum iteration number is 40k and we apply early stop at 5k iteration to reduce overfitting.
The GTA5 and BDD100K images are resized to $1280 \times 720$ and $960 \times 540$ for training respectively.
For the IBN-Net~\cite{pan2018two} baseline, we replace the batch normalization layers after the \{2, 4, 7\}-\textit{th} convolution layers with instance normalization layers.
For the AdaptSeg~\cite{tsai2018learning} baseline, we use the Adam optimizer~\cite{kingma2014adam} and 0.005 initial learning rate for the discriminator.

In our method, we use dynamic transferable embedding and curriculum training in addition to adversarial adaptation ~\cite{tsai2018learning}.
The visual memory here also comprises of a set of class centroids, which is an aggregation of local features belonging to the same category. 
Inspired by~\cite{zou2018unsupervised}, the curriculum learning procedure is further designed to include the averaged probability confidence of each image as a guidance.
Specifically, the samples that are easier, \textit{i.e.}, have higher confidence, are firstly fed into the model for adaptation.
Since domain encoder is not accessible here, we only use class enhancer for dynamic transferable embedding.
%
%

\noindent
\textbf{C-Mazes.}
The experiments for C-Mazes are also conducted in two-stages. In the first stage, We follow the setups in \cite{hu2018synthesized} to train a randomly initialized ResNet-18 policy network. We only use one single topology and two different tasks for the experiments. The total episodes is set to 16000. The initial learning rate is set to 0.001. Then in the second stage, we use the pre-trained model to calculate state feature centers for each actions and use memory module to fine-tune the model.

\section{Additional Results}
\label{sec:results}

\noindent
\textbf{C-Digits.}
We have experimented with mnist, mnist-m or usps as the source domain.  On average, the performance gain of our approach is 8.1\% over the baseline method JAN~\cite{long2017deep} and 8.9\% over MCD~\cite{saito2018maximum}; Likewise, the average performance gain with Multi-PIE as the source domain is 36.5\% and 14.4\% over the baselines.

\setlength{\tabcolsep}{4pt}
\begin{table*}[t!]
\caption{\textbf{Per-category IoU(\%) results on the C-Driving Benchmark.} (BDD100K dataset is used as the real-world target domain data.) The 'train' and 'bicycle' categories are not listed because their results are close to zero. }
\label{tab:driving_detail}
	\begin{center}
		\resizebox{17.0cm}{!}{
			\begin{tabular}{l|l|ccccccccccccccccc|c}
				\Xhline{2\arrayrulewidth}
				Domain & Method & \R{road} & \R{sidewalk} & \R{building} & \R{wall} & \R{fence} & \R{pole} & \R{light} & \R{sign} & \R{veg} & \R{terrain} & \R{sky} & \R{person} & \R{rider} & \R{car} & \R{truck} & \R{bus} & \R{mcycle} & mIoU \\ \hline\hline
				\multirow{5}{*}{Rainy} & Source only & 48.3 & 3.4 & 39.7 & 0.6 & 12.2 & 10.1 & 5.6 & 5.1 & 44.3 & 17.4 & 65.4 & 12.1 & 0.4 & 34.5 & 7.2 & 0.1 & 0.5 & 16.2 \\
				 & AdaptSeg~\cite{tsai2018learning} & 58.6 & 17.8 & 46.4 & 2.1 & 19.6 & 15.6 & 5.0 & 7.7 & \textbf{55.6} & \textbf{20.7} & 65.9 & 17.3 & 0.0 & 41.3 & 7.4 & 3.1 & 0.0 & 20.2 \\
				 & CBST~\cite{zou2018unsupervised} & 59.4 & 13.2 & 47.2 & 2.4 & 12.1 & 14.1 & 3.5 & 8.6 & 53.8 & 13.1 & \textbf{80.3} & 13.7 & \textbf{17.2} & \textbf{49.9} & 8.9 & 0.0 & \textbf{6.6} & 21.3 \\
				 & IBN-Net~\cite{pan2018two} & 58.1 & \textbf{19.5} & 51.0 & \textbf{4.3} & \textbf{16.9} & \textbf{18.8} & 4.6 & \textbf{9.2} & 44.5 & 11.0 & 69.9 & \textbf{20.0} & 0.0 & 39.9 & 8.4 & 15.3 & 0.0 & 20.6 \\
				 & Ours & \textbf{63.0} & 15.4 & \textbf{54.2} & 2.5 & 16.1 & 16.0 & \textbf{5.6} & 5.2 & 54.1 & 14.9 & 75.2 & 18.5 & 0.0 & 43.2 & \textbf{9.4} & \textbf{24.6} & 0.0 & \textbf{22.0} \\ \hline
				\multirow{5}{*}{Snowy} & Source only & 50.8 & 4.7 & 45.1 & \textbf{5.9} & \textbf{24.0} & 8.5 & 10.8 & 8.7 & 35.9 & 9.4 & 60.5 & 17.3 & 0.0 & 47.7 & 9.7 & 3.2 & \textbf{0.7} & 18.0 \\
				 & AdaptSeg~\cite{tsai2018learning} & 59.9 & 13.3 & 52.7 & 3.4 & 15.9 & 14.2 & 12.2 & 7.2 & \textbf{51.0} & \textbf{10.8} & 72.3 & 21.9 & 0.0 & 55.0 & 11.3 & 1.7 & 0.0 & 21.2 \\
				 & CBST~\cite{zou2018unsupervised} & 59.6 & 11.8 & 57.2 & 2.5 & 19.3 & 13.3 & 7.0 & \textbf{9.6} & 41.9 & 7.3 & 70.5 & 18.5 & 0.0 & \textbf{61.7} & 8.7 & 1.8 & 0.2 & 20.6 \\
				 & IBN-Net~\cite{pan2018two} & 61.3 & \textbf{13.5} & 57.6 & 3.3 & 14.8 & \textbf{17.7} & 10.9 & 6.8 & 39.0 & 6.9 & 71.6 & \textbf{22.6} & 0.0 & 56.1 & \textbf{13.8} & \textbf{20.4} & 0.0 & 21.9 \\
				 & Ours & \textbf{68.0} & 10.9 & \textbf{61.0} & 2.3 & 23.4 & 15.8 & \textbf{12.3} & 6.9 & 48.1 & 9.9 & \textbf{74.3} & 19.5 & 0.0 & 58.7 & 10.0 & 13.8 & 0.1 & \textbf{22.9} \\ \hline
				\multirow{5}{*}{Cloudy} & Source only & 47.0 & 8.8 & 33.6 & 4.5 & 20.6 & 11.4 & \textbf{13.5} & 8.8 & 55.4 & 25.2 & 78.9 & 20.3 & 0.0 & 53.3 & 10.7 & 4.6 & 0.0 & 20.9 \\
				 & AdaptSeg~\cite{tsai2018learning} & 51.8 & 15.7 & 46.0 & 5.4 & \textbf{25.8} & 18.0 & 12.0 & 6.4 & \textbf{64.4} & 26.4 & 82.9 & \textbf{24.9} & 0.0 & 58.4 & 10.5 & 4.4 & 0.0 & 23.8 \\
				 & CBST~\cite{zou2018unsupervised} & 56.8 & \textbf{21.5} & 45.9 & 5.7 & 19.5 & 17.2 & 10.3 & 8.6 & 62.2 & 24.3 & \textbf{89.4} & 20.0 & 0.0 & 58.0 & \textbf{14.6} & 0.1 & 0.1 & 23.9 \\
				 & IBN-Net~\cite{pan2018two} & 60.8 & 18.1 & 50.5 & \textbf{8.2} & 25.6 & \textbf{20.4} & 12.0 & \textbf{11.3} & 59.3 & 24.7 & 84.8 & 24.1 & \textbf{12.1} & 59.3 & 13.7 & 9.0 & 1.2 & 26.1 \\
				 & Ours & \textbf{69.3} & 20.1 & \textbf{55.3} & 7.3 & 24.2 & 18.3 & 12.0 & 7.9 & 64.2 & \textbf{27.4} & 88.2 & 24.7 & 0.0 & \textbf{62.8} & 13.6 & \textbf{18.2} & 0.0 & \textbf{27.0} \\ \hline
				\multirow{5}{*}{Overcast} & Source only & 46.6 & 9.5 & 38.5 & 2.7 & 19.8 & 12.9 & \textbf{9.2} & 17.5 & 52.7 & 19.9 & 76.8 & 20.9 & 1.4 & 53.8 & 10.8 & 8.4 & \textbf{1.8} & 21.2 \\
				 & AdaptSeg~\cite{tsai2018learning} & 59.5 & 24.0 & 49.4 & 6.3 & 23.3 & 19.8 & 8.0 & 14.4 & \textbf{61.5} & 22.9 & 74.8 & 29.9 & 0.3 & 59.8 & 12.8 & 9.7 & 0.0 & 25.1 \\
				 & CBST~\cite{zou2018unsupervised} & 58.9 & \textbf{26.8} & 51.6 & 6.5 & 17.8 & 17.9 & 5.9 & \textbf{17.9} & 60.9 & 21.7 & \textbf{87.9} & 22.9 & 0.0 & 59.9 & 11.0 & 2.1 & 0.2 & 24.7 \\
				 & IBN-Net~\cite{pan2018two} & 62.9 & 25.3 & 55.5 & 6.5 & 21.2 & \textbf{22.3} & 7.2 & 15.3 & 53.3 & 16.5 & 81.6 & \textbf{31.1} & \textbf{2.4} & 59.1 & 10.3 & \textbf{14.2} & 0.0 & 25.5 \\
				 & Ours & \textbf{73.5} & 26.5 & \textbf{62.5} & \textbf{8.6} & \textbf{24.2} & 20.2 & 8.5 & 15.2 & 61.2 & \textbf{23.0} & 86.3 & 27.3 & 0.0 & \textbf{64.4} & \textbf{14.3} & 13.3 & 0.0 & \textbf{27.9} \\ \Xhline{2\arrayrulewidth}
			\end{tabular}
		}
	\end{center}
\end{table*}
\setlength{\tabcolsep}{1.4pt}

\begin{figure*}[t]
  \centering
  \includegraphics[width=0.8\textwidth]{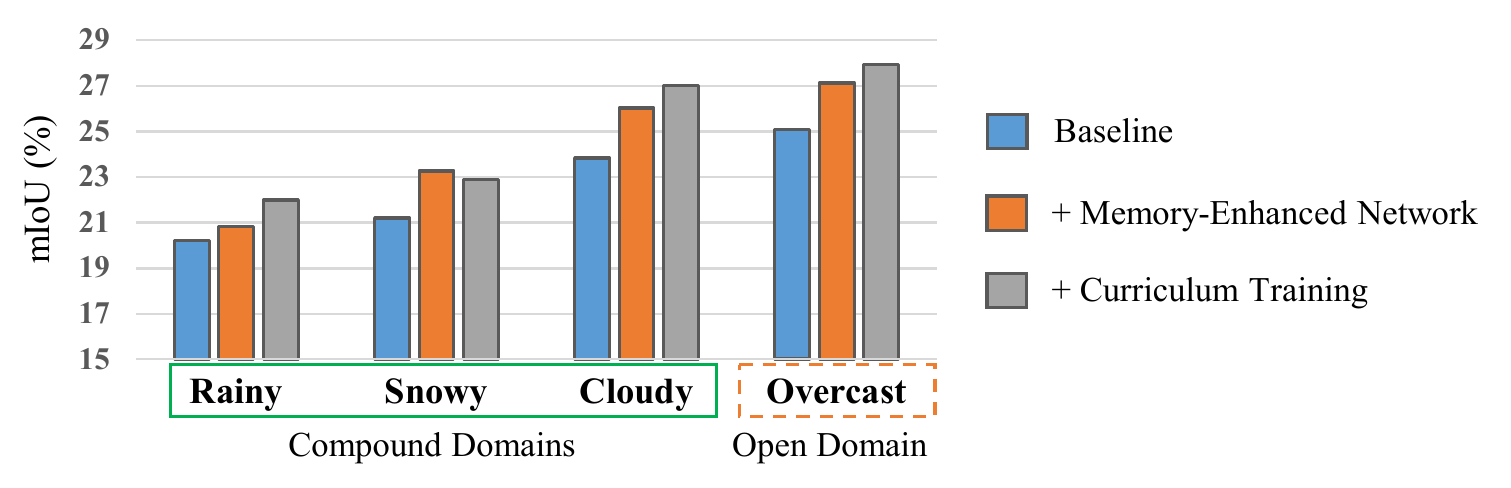}
  \caption{\textbf{Ablation study} of memory-enhanced neural network and curriculum training on the C-Driving benchmark (GTA5 and BDD100K datasets).}
  \label{fig:ablation_drive}
\end{figure*}

\begin{figure*}[t]
  \centering
  \includegraphics[width=1.0\textwidth]{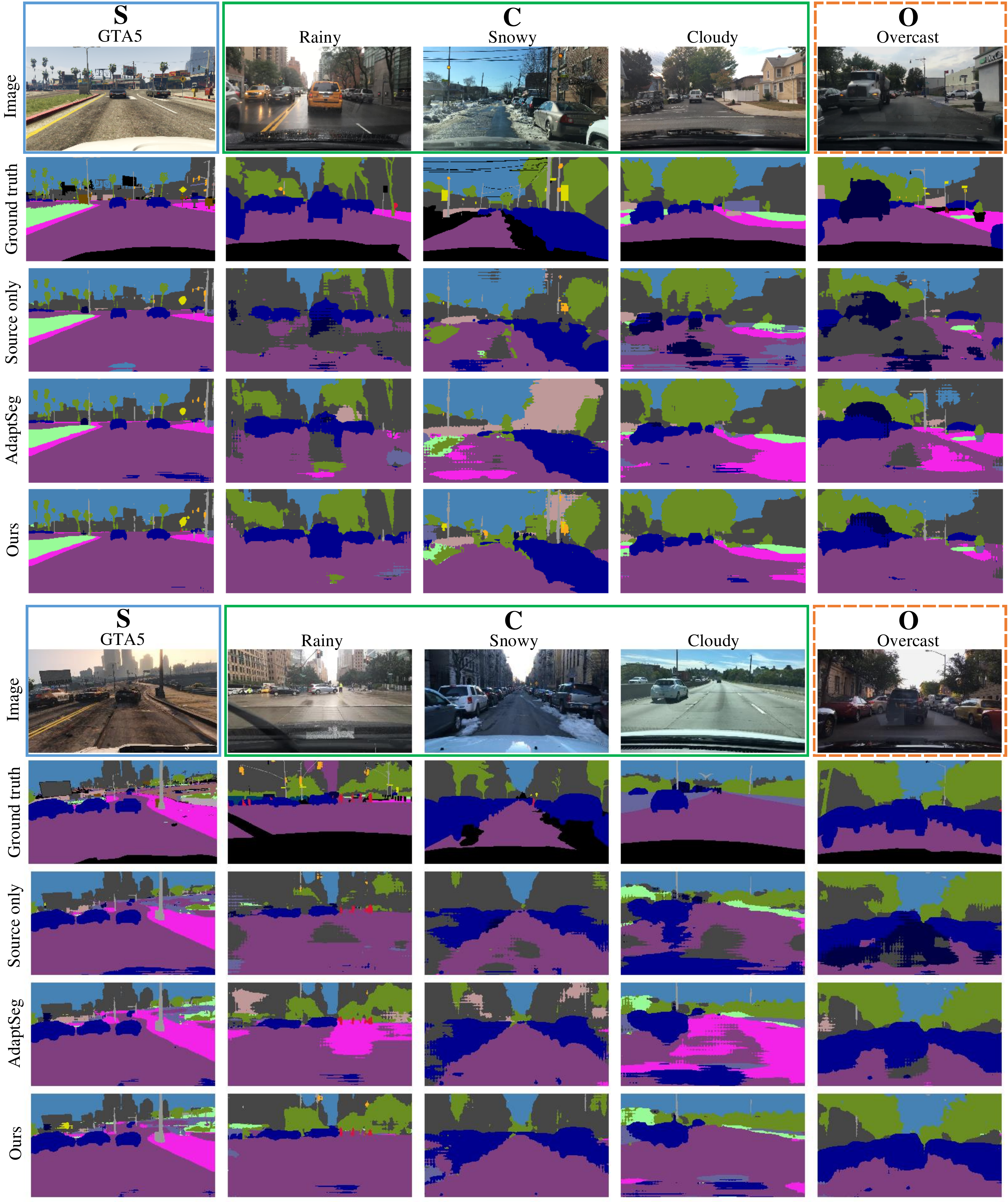}
  \caption{\textbf{Qualitative results comparison of semantic segmentation} on the source domain (\textbf{S}), the compound domains (\textbf{C}), and the open domain (\textbf{O}).}
  \label{fig:visualize_drive}
\end{figure*}

\noindent
\textbf{C-Driving.}
Figure \ref{fig:visualize_drive} shows example results on the GTA5 and BDD100K dataset.
Our method produces more accurate segmentation results on the compound domains and the open domain compared to 'source only' and AdaptSeg~\cite{tsai2018learning}.
Per-category results are provided in Table~\ref{tab:driving_detail}, and ablation study of dynamic transferable embedding and curriculum training are shown in Figure~\ref{fig:ablation_drive}.

\noindent
\textbf{Office-Home~\cite{venkateswara2017deep}.}
We had some preliminary results on Office-Home~\cite{venkateswara2017deep}, where our approach outperforms baseline methods (JAN~\cite{long2017deep} and MCD~\cite{saito2018maximum}) by 15.3\% and 7.9\%, respectively.



\section{More Visualizations}
\label{sec:visualization}

\begin{figure*}[t]
  \centering
  \includegraphics[width=1.0\textwidth]{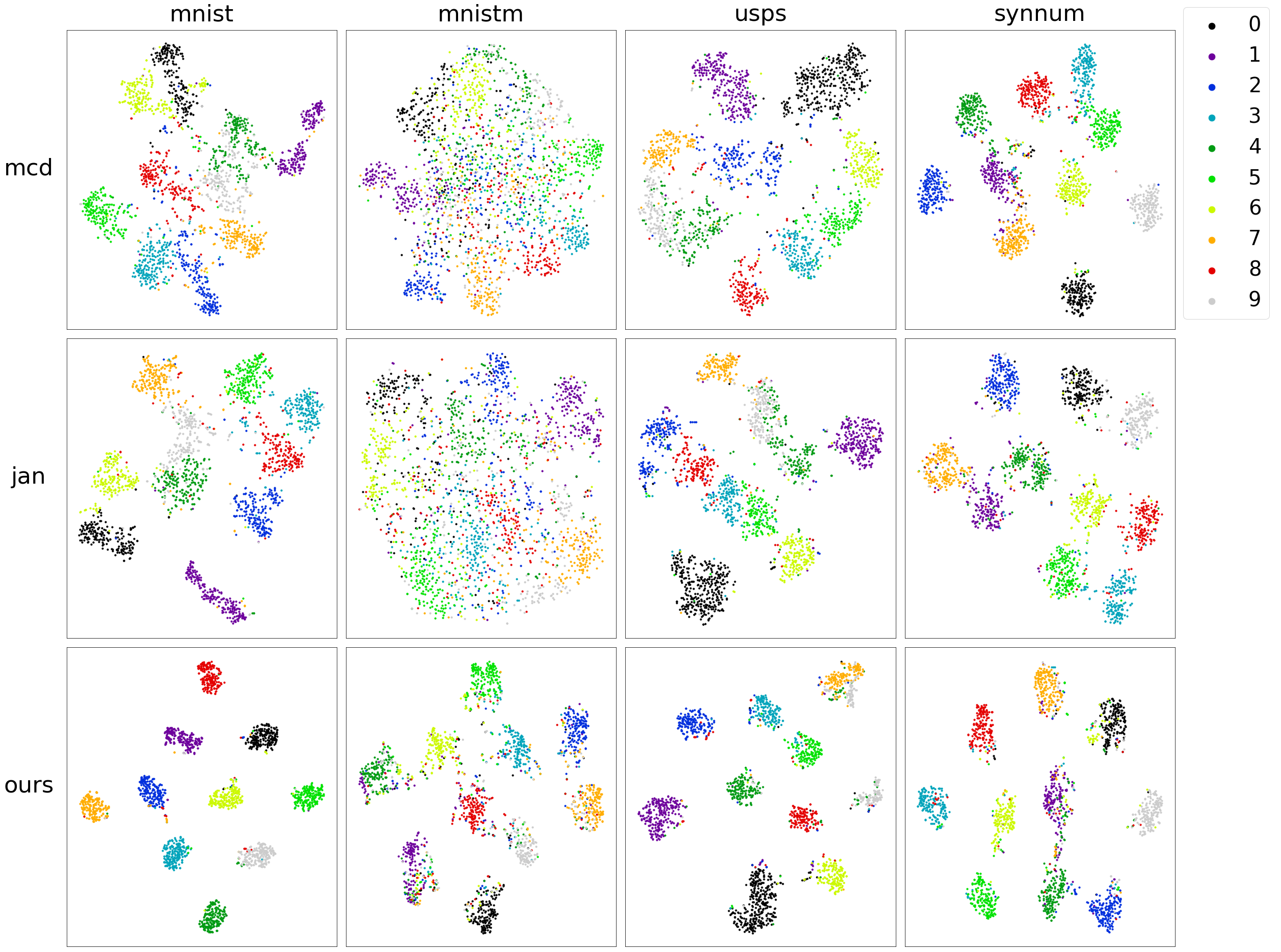}
  \caption{\textbf{t-SNE of the C-Digits features} of MCD~\cite{saito2018maximum}, JAN~\cite{long2017deep}, and our approach.}
  \label{fig:tsne-digits}
\end{figure*}

\begin{figure*}[t]
  \centering
  \includegraphics[width=1.0\textwidth]{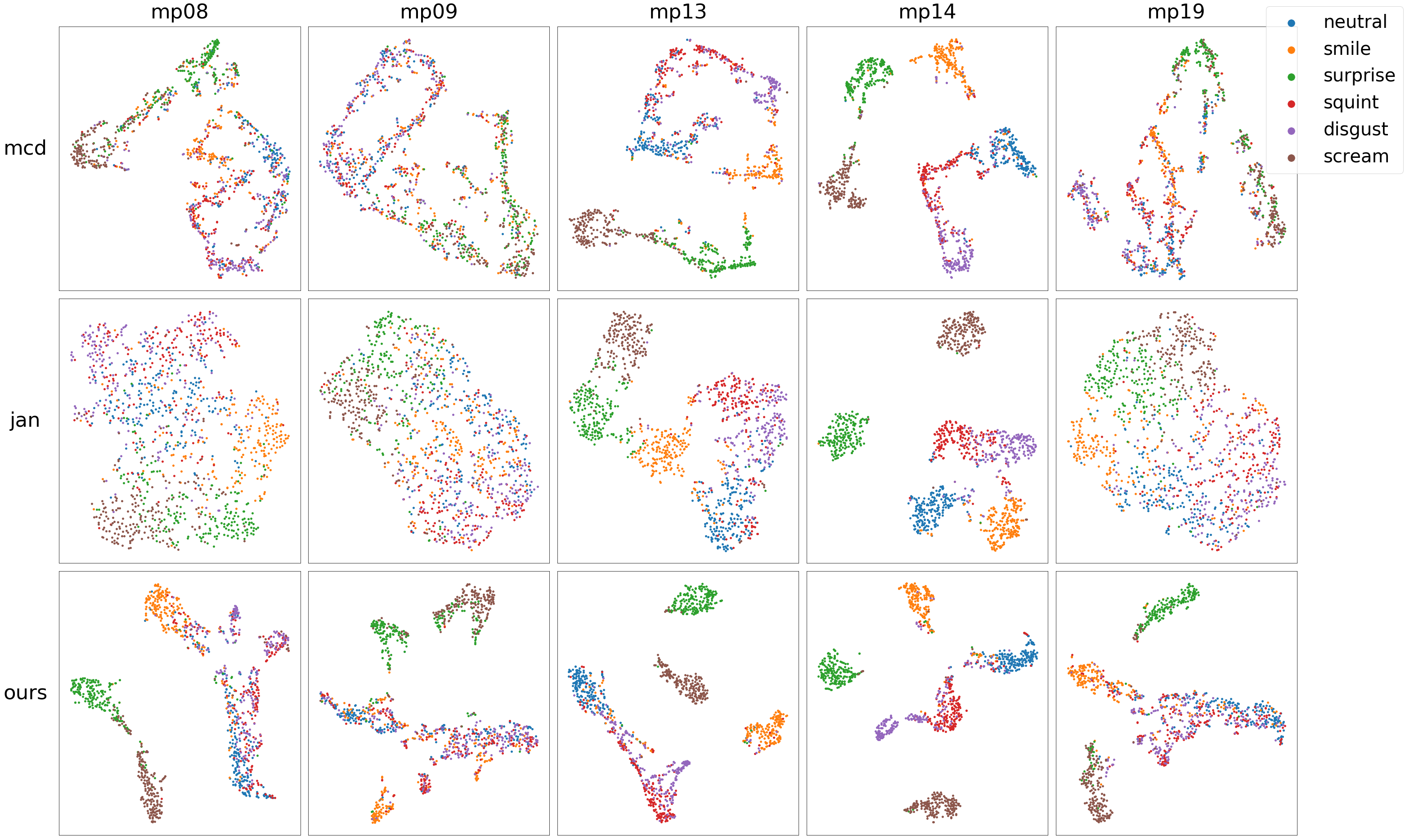}
  \caption{\textbf{t-SNE of the C-Faces expression features} of MCD~\cite{saito2018maximum}, JAN~\cite{long2017deep}, and our approach.}
  \label{fig:tsne-face}
\end{figure*}

\begin{figure*}[t]
  \centering
  \includegraphics[width=1.0\textwidth]{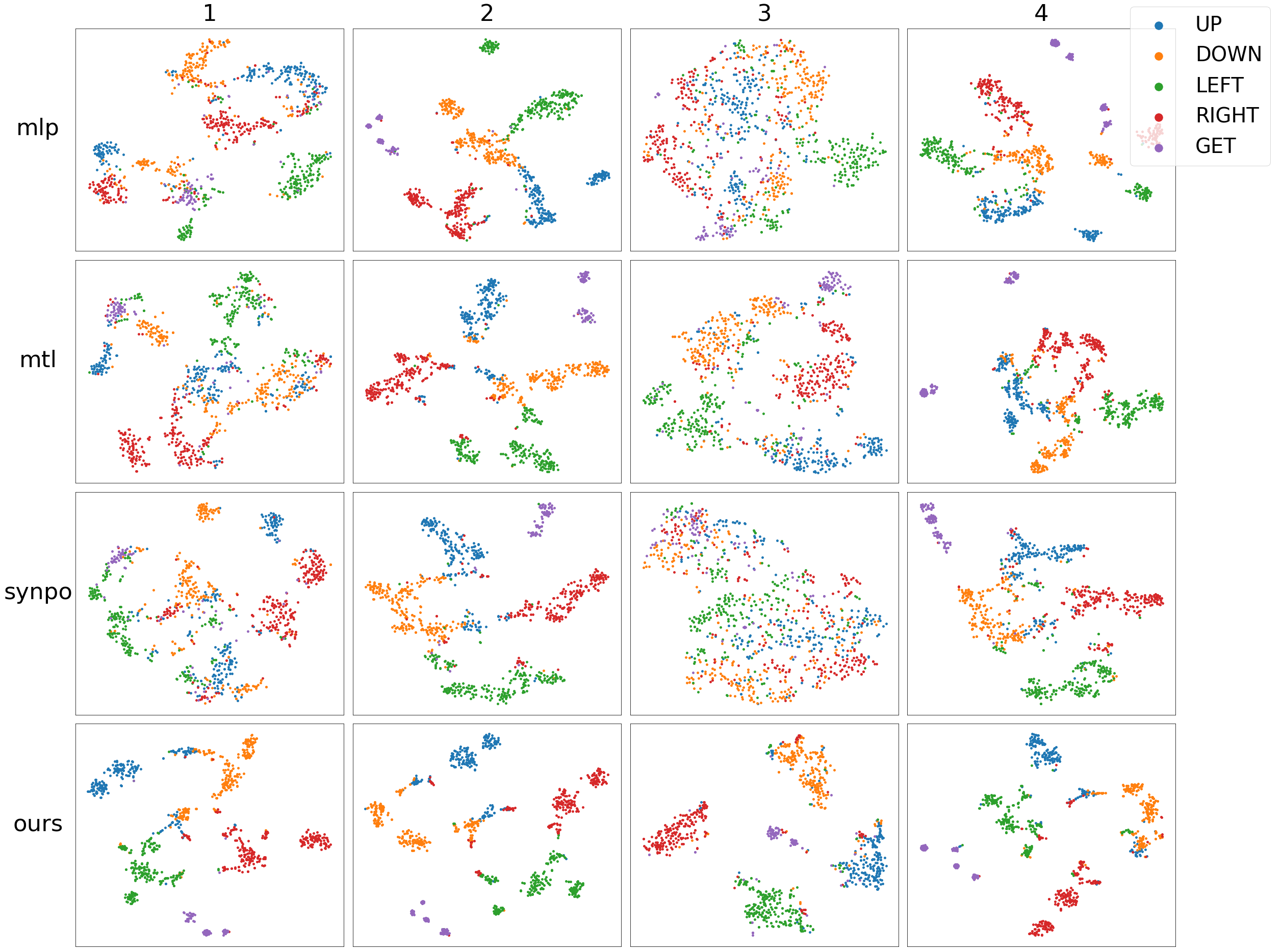}
  \caption{\textbf{t-SNE of the C-Maze action features} of MLP~\cite{hu2018synthesized}, MTL, SynPo~\cite{hu2018synthesized}, and our approach.}
  \label{fig:tsne-maze}
\end{figure*}

\noindent
\textbf{t-SNE Visualization.}
Here we show t-SNE visualizations of the learned dynamic transferable embedding on the C-Digits, C-Faces, and C-Mazes testing data (Figure~\ref{fig:tsne-digits} - \ref{fig:tsne-maze}), among various methods. The dynamic transferable embedding on the C-Mazes benchmark are state features of each actions. Our approach generally learns a more discriminative feature space thanks to the proposed disentanglement and memory modules.

\begin{figure*}[t]
  \centering
  \includegraphics[width=1.0\textwidth]{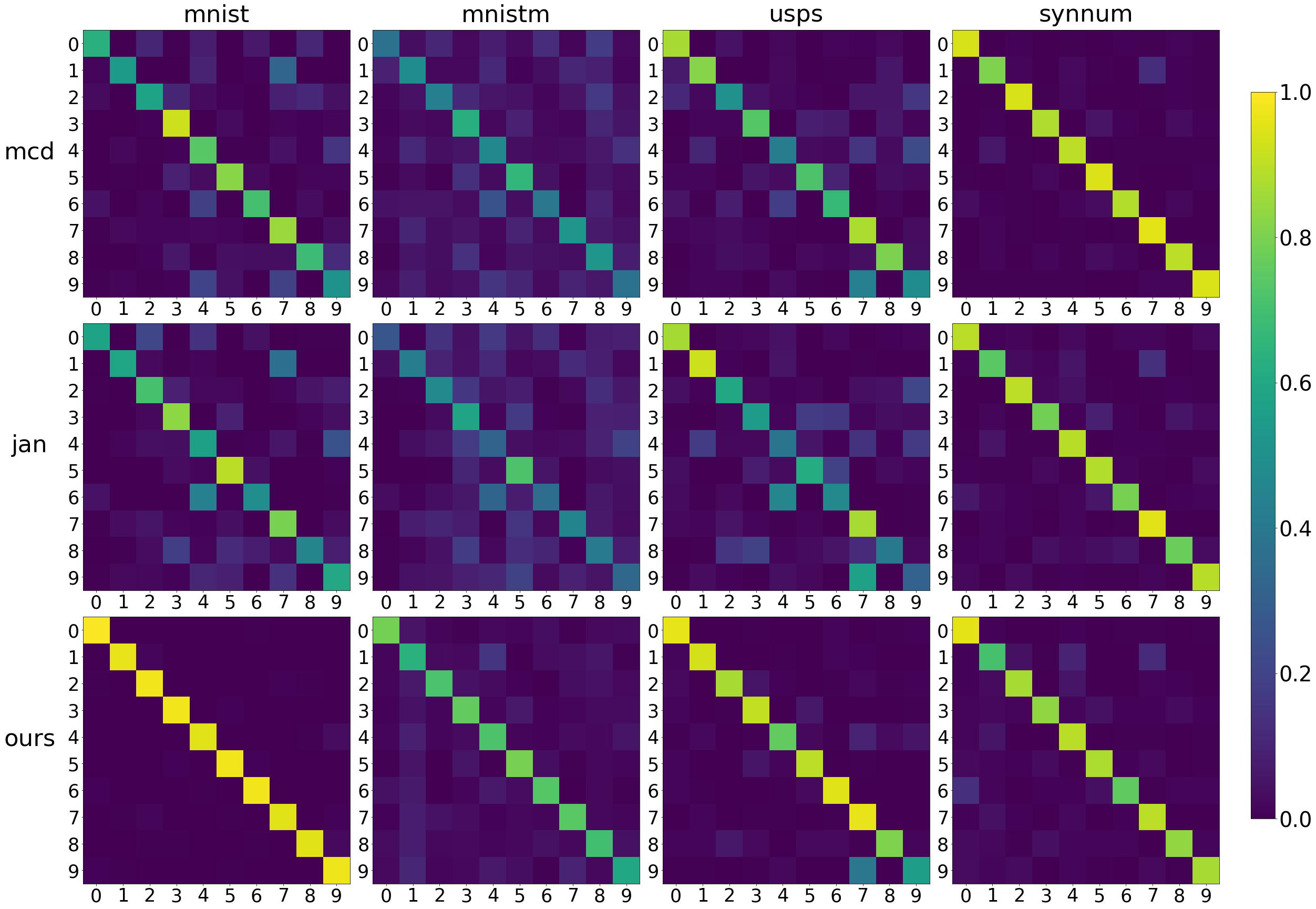}
  \caption{\textbf{Confusion matrices visualizations on the C-Digits benchmark} for MCD~\cite{saito2018maximum}, JAN~\cite{long2017deep}, and our approach.}
  \label{fig:conf-digits}
\end{figure*}

\begin{figure*}[!htp]
  \centering
  \includegraphics[width=1.0\textwidth]{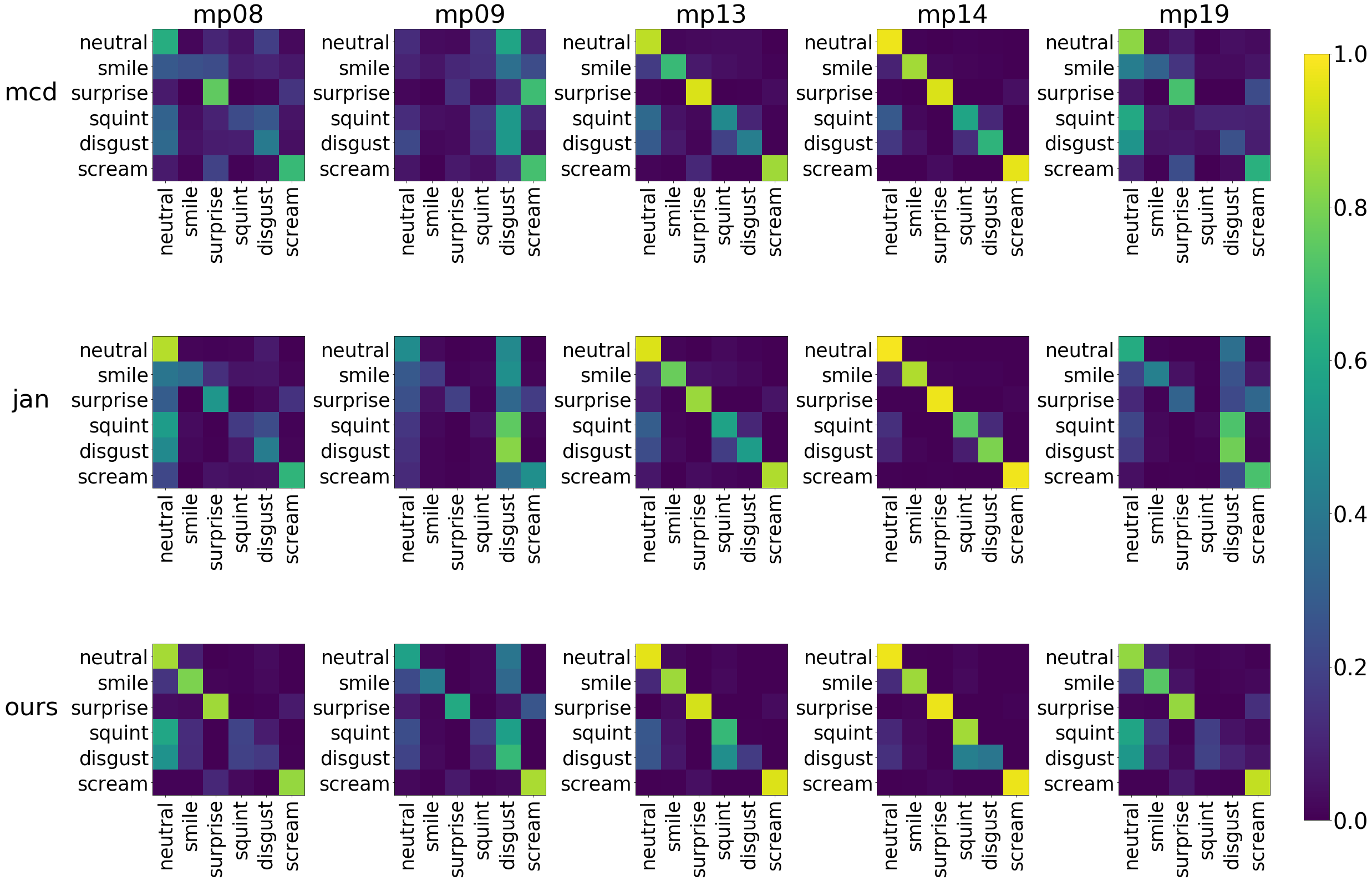}
  \caption{\textbf{Confusion matrices visualizations on the C-Faces benchmark} for MCD~\cite{saito2018maximum}, JAN~\cite{long2017deep}, and our approach.}
  \label{fig:conf-faces}
\end{figure*}

\noindent
\textbf{Confusion Matrices.}
Here we show visualizations of class confusion matrices on the C-Digits and C-Faces testing data in Figure~\ref{fig:conf-digits} and Figure~\ref{fig:conf-faces}. Compared to MCD~\cite{saito2018maximum} and JAN~\cite{long2017deep}, our approach performs better on the discriminative accuracies of each class.

\end{document}